\documentclass[letterpaper, 10 pt, conference]{ieeeconf}
\IEEEoverridecommandlockouts
\usepackage[utf8]{inputenc}
\usepackage{pgfplots}
\DeclareUnicodeCharacter{2212}{−}
\usepgfplotslibrary{groupplots,dateplot}
\usetikzlibrary{patterns,shapes.arrows}
\pgfplotsset{compat=newest}

\usepackage{xspace}
\usepackage{amsmath}
\usepackage{amssymb}
\usepackage{mathtools}
\usepackage{nicefrac}
\usepackage{amsfonts}
\usepackage{pifont}
\usepackage{floatflt}
\usepackage{url}
\usepackage{dsfont}

\usepackage{tikz}
\usetikzlibrary{calc, positioning, shadows, backgrounds, arrows.meta}
\usepackage{multirow}
\usepackage{booktabs}
\usepackage{tabularx}
\newcolumntype{L}{>{\raggedright\arraybackslash}X}

\usepackage{algorithm}

\makeatletter
\DeclareRobustCommand*\cal{\@fontswitch\relax\mathcal}
\makeatother

\newcommand{\xai}{\textsc{XAI}\xspace}

\newcommand{\commentout}[1]{}

\newcommand{\V}{{\cal V}}

\newcommand{\sat}{\models}

\newcommand{\lem}{\begin{lemma}}
\newcommand{\elem}{\end{lemma}}
\newcommand{\pro}{\begin{proposition}}
\newcommand{\epro}{\end{proposition}}

\newcommand{\dfn}{\begin{definition}}
\newcommand{\edfn}{\end{definition}}

\newcommand{\xam}{\begin{example}}
\newcommand{\exam}{\end{example}}

\usepackage{algorithm}

\usepackage{longtable}
\usepackage{subcaption}
\usepackage{pgfplots}
\usepackage{pgfplotstable}
\usepackage{paralist}
\usepgfplotslibrary{fillbetween}
\pgfplotsset{compat=1.16}

\newcommand{\?}{\stackrel{?}{=}}

\usepackage{comment}
\usepackage{graphicx}
\usepackage{hyperref}
\usepackage{xcolor}
\let\labelindent\relax
\usepackage{enumitem}
\usepackage[noend]{algpseudocode}
\algrenewcommand\algorithmicindent{0.6em}%
\usepackage[capitalize,noabbrev]{cleveref}
\usepackage{pgfplots}
\pgfplotsset{compat=1.18}
\usepackage{tikz}
\usetikzlibrary{fit}
\usepackage{xspace}
\usepackage{multirow}
\usepackage{orcidlink}

\crefname{line}{line}{lines}
\crefname{figure}{Fig.}{Figs.}
\Crefname{figure}{Fig.}{Figs.}
\crefname{equation}{Eq.}{Eqs.}
\Crefname{equation}{Eq.}{Eqs.}
\crefname{section}{Sec.}{Secs.}
\Crefname{section}{Sec.}{Secs.}
\crefname{definition}{Def.}{Defs.}
\Crefname{definition}{Def.}{Defs.}
\crefname{algorithm}{Alg.}{Algs.}
\Crefname{algorithm}{Alg.}{Algs.}
\Crefname{algocf}{Alg.}{Algs.}
\Crefname{appendix}{Appendix}{Appendices}

\newtheorem{example}{Example}
\newtheorem{lemma}{Lemma}

\newtheorem{definition}{Definition}

\newtheorem{proposition}{Proposition}
\newtheorem{observation}{Observation}

\newcommand{\status}{\mathtt{status}}

\newcommand{\state}{\xi}
\newcommand{\statetraj}{\overline{\state}}

\newcommand{\env}{\textit{env}}

\newcommand{\aref}[1]{\hyperref[#1]{Appendix~\ref*{#1}}}

\newcommand{\KEnote}[1]{}

\DeclareMathOperator{\ac}{ac}

\begin{document}
\graphicspath{{./images/}}

\title{\LARGE \bf
Causal Analysis of Failures of Cyber-Physical Systems\\with Continuous Inputs
}


\author{
Tom Yaacov$^{*,1}$\orcidlink{0000-0002-0565-6506}, David A. Kelly$^{1}$\orcidlink{0000-0002-5368-6769}, Khen Elimelech$^{1}$\orcidlink{0000-0003-1391-8956}, and Hana Chockler$^{1}$\orcidlink{0000-0003-1219-0713} 
\thanks{$^{1}$ King's College London,
        {\tt\footnotesize \{firstname.lastname\}@kcl.ac.uk}}%
\thanks{* Corresponding Author.}
\thanks{** This work was partially supported by the Causality in Healthcare AI (CHAI) Hub [UKRI AI and EPSRC grant EP/Y028856/1].}
}

\maketitle

\begin{abstract}
Modern autonomous Cyber-Physical Systems (CPSs), such as self-driving cars, operate in increasingly complex environments while expected to maintain high reliability. The black-box nature of many of these systems makes it impossible to fully verify their behavior prior to deployment. When a failure occurs, it is important to analyze it to guide improvements to system robustness. A recent work introduced the novel concept of leveraging the framework of actual causality for CPS failure explanation. However, that work considered only a discrete, binary causal model that cannot capture continuous changes in system inputs. In this paper, we address these limitations by extending the actual causality framework with the notion of distance to account for the continuous characteristics of the system. Specifically, we introduce the novel concepts of distance-weighted actual cause and distance-weighted degree of responsibility, enabling causal influence to account for the magnitude of changes required to avoid a failure, rather than only its cardinality. While motivated by the continuous nature of CPSs, this constitutes a general extension of the theory of actual causality and enables more fine-grained analysis of continuous inputs in any causal model. Building on these notions, we present an efficient, practical, system-agnostic explanation-derivation algorithm that uses distance-weighted responsibility to guide the explanation search. We experimentally evaluate our approach across a number of environments from two autonomous-driving simulators with black-box neural controllers. The results show that the proposed extension provides more effective guidance than the current discrete approach.

\end{abstract}

\section{Introduction}

Autonomous robots and Cyber-Physical Systems (CPSs) are becoming increasingly prevalent in modern life. Such systems leverage digital components to control physical machines that operate in and interact with the real world. To operate safely and reliably, these systems must be robust, i.e., resist failures and maintain performance under a wide range of circumstances~\cite{10.1145/3170500,10.3389/fenrg.2023.1095303}. However, due to the complexity of modern systems and the intractability of real-world scenarios, verifying their robustness prior to deployment is unrealistic. This is especially true for systems built on ``black-box'' neural components, such as controllers and perception units, which typically can only be verified empirically.

When a failure occurs, it is important to analyze it to help ensure that similar scenarios do not repeat in the future, or at least to ensure the user is better prepared to mitigate them if they do. 
Perhaps the most useful output of this analysis is an \emph{explanation}, 
identifying the elements in the potentially rich scenario that led to a failure.

\textbf{Problem definition:}
\textbf{Our goal is, given a specific failure example (input environment and simulated system run), to derive an explanation of the cause of failure. 
We aim to derive a valid explanation while minimizing the computational effort required to find it.}

While the usefulness of causal analysis has been demonstrated for discrete systems, there is little work to connect it to CPS, which have challenging properties such as continuous input variables and temporally extended operations. A recent work~\cite{icra} was the first to leverage the framework of \emph{actual causality} to explain CPS failures. We note that, unlike the more common framework of general causality~\cite{Pearl2009causality}, which derives forward-looking, probabilistic causal models, actual causality~\cite{Hal16} is \emph{backward-looking}, using causal models to explain specific past events. The work in~\cite{icra} took the first step toward bridging the gap between CPSs and actual causality, providing guidelines for integrating a system-testing model with a causal model to ensure correct failure explanations. Since the standard search for such explanations quickly becomes intractable, it also introduced an efficient heuristic-guided explanation-derivation algorithm based on causal \emph{degree of responsibility}~\cite{CH04,Hal16}---a quantification of the causal influence of an input variable.

While serving as a stepping stone, that work made significant assumptions. Most importantly, it did not allow for continuous variability of the testing-model inputs. This is a critical limitation, as explanation derivation requires examining modified scenarios and reasoning about their effect on system behavior. Yet, the causal-model inputs accepted only binary valuations: elements could either exist or not exist, but could not vary continuously. This assumption clashes with the continuous nature of CPSs and potentially limits the utility and accuracy of explanations that can be derived. Moreover, this limitation carries over to responsibility, the key heuristic guiding the explanation search. When evaluating modified scenarios to compute an element's responsibility, the approach cannot distinguish between small and large modifications, as all modifications are treated equally. Consequently, the responsibility measure loses much of its nuance, providing less effective guidance for the search.


In this work, we address these limitations by \textbf{extending actual causality to account for the continuous characteristics of CPSs} and exploiting this extension to \textbf{effectively derive causal explanations of system failures}. In particular, we extend actual causality with the notion of distance, enabling a more fine-grained analysis of continuous inputs. While motivated by the continuous nature of CPSs, this extension constitutes a \textbf{novel contribution to the theory of actual causality} in general, allowing causal influence to account for the magnitude of interventions (i.e., changes to input variables) rather than only their presence. We then leverage this extension and propose an improved responsibility-guided search approach for failure explanations.

Specifically, we make the following contributions: \textbf{(1)}~We \textbf{extend the framework for CPS failure explanation} presented in~\cite{icra} to support continuous causal models. \textbf{(2)}~We \textbf{extend the framework of actual causality itself with the notions of distance-weighted actual cause and distance-weighted responsibility}, allowing causal analysis to account for the magnitude of interventions. \textbf{(3)}~Building on these notions, we present an \textbf{improved and more efficient explanation-derivation algorithm for black-box CPSs} that uses distance-weighted responsibility to guide the explanation search. \textbf{(4)}~We experimentally evaluate our approach against prior work on autonomous-driving systems, demonstrating that it provides \textbf{more effective and scalable guidance for explanation search}. 
We further evaluate distance-weighted responsibility by assessing \textbf{how well it prioritizes the elements involved across multiple possible explanations for a failure}, providing further evidence of its effectiveness.




The paper is organized as follows. \cref{sec:prelim}~provides preliminaries on the CPS testing model and the actual causality framework and formulates our objectives. \cref{sec:extending-ac}~describes the extension of the actual causality framework with the notion of distance. \cref{sec:method}~presents the distance-responsibility-guided failure explanation generation algorithm. The experimental evaluation is covered in \cref{sec:evaluation}, and \cref{sec:conclusion}~concludes the paper.

\noindent\textbf{Related Work.}
Actual causality~\cite{Hal16,CH24} is well-suited for explaining events that have already occurred and has been extensively applied in explainable AI (\xai), including image, audio and medical classification~\cite{CH24,CKKS26,kelly2026guess,blake2026causal}. The \xai tool \textsc{ReX} approximates causal responsibility to efficiently derive causal explanations~\cite{CKK25}. These approaches, like ours, treat the AI model as a black box. 
Notably,~\cite{CKKS26} also uses causal responsibility~\cite{CH04} for guiding explanation generation, but is specifically designed for image classification.

In the context of CPSs, the most relevant work to this paper is~\cite{icra}, which uses actual causality to explain failures. 
Its algorithm is based on causal responsibility, however, it treats all variables as binary, hence  
cannot capture or manipulate continuous dependencies between inputs and outcomes. 
Counterfactual reasoning has recently been used for CPS failure analysis~\cite{ghazal2026towards}. 
Actual causality extends and generalizes counterfactual reasoning, allowing to capture more subtle dependencies 
~\cite{Hal16}.

Sensitivity analysis, described in~\cite{diwakaran2017}, perturbs the position of obstacles in failure-causing scenes to analyze CPSs. This analysis
can identify parts of the environment associated with the failure, but does not generate explanations.
\cite{bartocci} uses a debugger to generate explanations of instrumented CPS failures, an approach whose feasibility depends strongly on the model. 
In contrast, our method treats the system as a black box. 
\cite{diehl2022} learns a Bayesian model from simulation data and uses it to generate explanations, a fundamentally different approach from actual causality. 
\cite{tenorth2009knowrob} and~\cite{lim2011ontology} construct ontology-based representations that integrate sensor and environment information with higher-level knowledge for robotic reasoning. Unlike our approach, these methods rely on explicit knowledge representations and reasoning mechanisms tailored to the robotic system. \cite{banerjee2021} compares system input/output traces using a state-machine encoding of the CPS rather than performing causal analysis; such a state machine may be unavailable or too complex to explain.

\begin{figure}[t]
\centering

\includegraphics[width=\columnwidth]{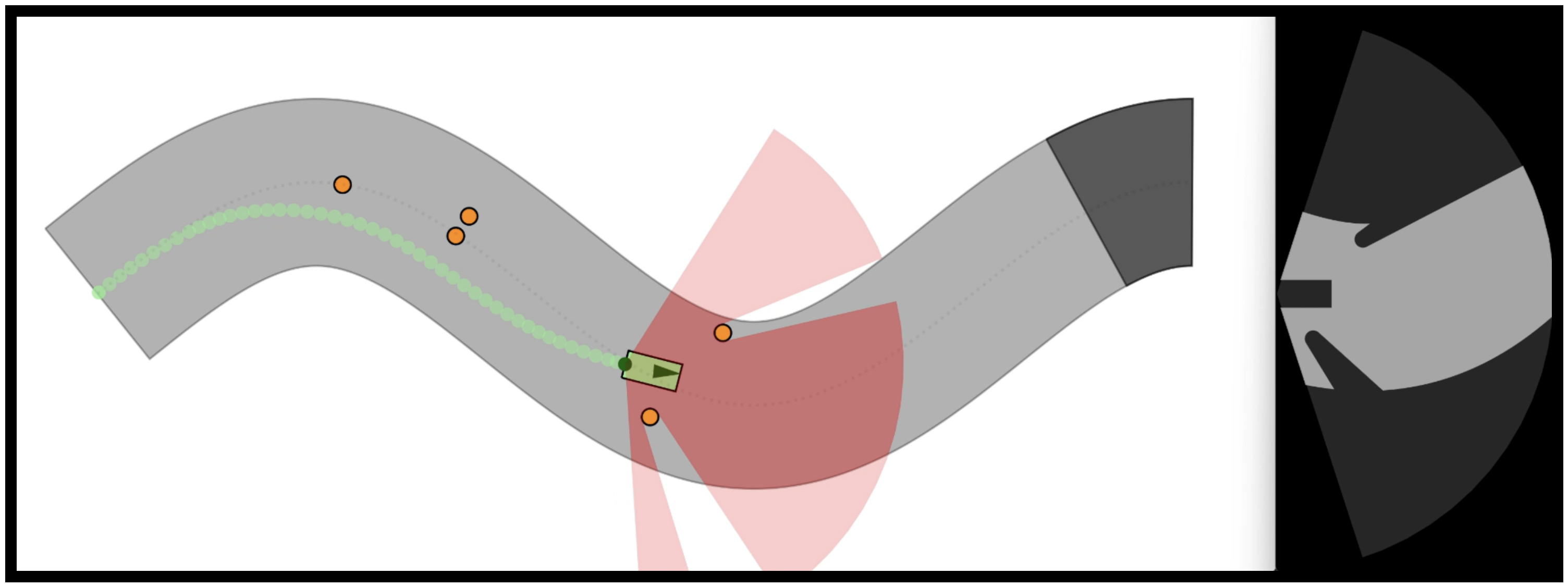}

\caption{The running example of an autonomous car, taken from the LiteRacer simulator~\cite{Elimelech24LiteRacer}. The car is controlled by a 
black-box neural network trained to navigate an obstructed track, avoiding collisions and reaching the end of the track.
}
\label{fig:car}
\end{figure}

\section{Preliminaries}
\label{sec:prelim}

\subsection{Testing Cyber-Physical Systems}

To facilitate the explanation of failures in autonomous CPSs, we adopt the formalism introduced in~\cite{elimelech2024falsification} and later extended in~\cite{icra}. We consider a CPS operating in an environment~$\env$, which encodes the external factors affecting the system's operation and specifies the variables describing its surroundings. The environment can be observed by the system and potentially changed by it. We assume that environment variability is the only source of variability in the system's execution, with no other external sources of uncertainty. We use $\state_t$ to denote the system state at time $t$, and $\statetraj$ to denote the resulting time-parameterized trajectory from start to finish. While a system execution is generally determined by both the environment $\env$ and the initial state $\state_0$, throughout this paper we assume that the initial state is fixed and therefore consider only the environment as input.

We assume that the CPS controller is designed to complete a specific \emph{task} while remaining robust to environmental variations. The outcome of an execution is determined by whether the resulting trajectory satisfies this task. Following~\cite{icra}, we distinguish not only between successful and failed executions but also between different types of failures. We capture this using a status predicate:
$$\status(\statetraj,\env) \in \{1, -1, -2, -3, \dots\},$$
where $\status(\statetraj,\env)=1$ indicates successful task completion, while each negative value indicates a different type of user-defined failure.



\subsection{Illustration: an autonomous car}
\label{sec:running-example}

While the ideas presented here apply to any CPS, we use an autonomous car operating on an obstructed track as a running example, as depicted in \cref{fig:car}. Autonomous vehicles are a representative example of CPSs, as their computational controllers process information from the physical environment and control the vehicle's behavior accordingly. In this example, variables define the track shape and the locations of circular obstacles. For simplicity, we assume the track shape is fixed and focus only on the obstacles. Each obstacle is represented by a vector $[x,y] \in [0,1]^2$, specified relative to the track: $x$ indicates its lateral position between the right and left shoulders, while $y$ indicates its longitudinal position from the starting line to the endzone. The state $\state_t \doteq [x_t,y_t,\phi_t,\alpha_t,v_t]$ consists of the car's position $(x_t,y_t)$, heading $\phi_t$, steering angle $\alpha_t$, and speed $v_t$, all in time $t$. The black-box controller is trained to steer the car to the end of the track while avoiding collisions, regardless of the track curvature and obstacle placement. We distinguish between failure types according to the location of the obstacle the car collides with. That is, the $\status$ predicate returns $1$ if the car successfully reaches the end of the track without a collision, and $-i$ if it crashes on or close
to the $i$-th obstacle.

\begin{figure}[t]
\centering
    \begin{tikzpicture}
    \node (V) at (-5, 0) {$\vec{V}:$};
    \node (v1) at (-3.5, 0) [draw, rectangle, rounded corners, minimum size=0.3cm,fill=red!20] {$(x_{v_1},y_{v_1})$};
    \node (v2) at (-1.5, 0) [draw, rectangle, rounded corners, minimum size=0.3cm,fill=red!20] {$(x_{v_2},y_{v_2})$};
    \node (dots) at (0, 0) {$\mathbf{\cdots}$};
    \node (vn) at (1.5, 0) [draw, rectangle, rounded corners, minimum size=0.3cm,fill=red!20] {$(x_{v_n},y_{v_n})$};

    \node (f) at (-1, -1.0) [draw, rectangle, rounded corners, fill=blue!20, minimum size=0.3cm]{$\status(\statetraj,\env) \? \status(\statetraj',\env')$};

    \draw [-Triangle] (v1) -- (f);
    \draw [-Triangle] (v2) -- (f);
    \draw [-Triangle] (vn) -- (f);
    
    \node (o) at (-1, -2.0) [draw, circle, minimum size=0.5cm, fill=red!20] {$O$};
    \node (oin) at (0.5, -2.0) {$O \in \{0,1\}$};

    \draw [-Triangle] (f) -- (o);
    \end{tikzpicture}
    \caption{A graphical representation of the causal model used to explain failures of our exemplary autonomous-car system. Unlike \cite{icra}, here the test inputs, i.e., obstacle placements, are modeled as \emph{continuous} location variables, rather than binary present/absent indicators. The outcome variable $O$ indicates whether the outcome of some execution under the intervention, $(\statetraj',\env')$, matches that of the original $(\statetraj,\env)$.}
    \label{fig:causal_model}
\end{figure}
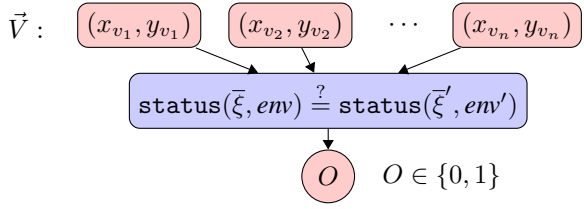

\subsection{Actual causality} 
\label{subsec:cause} 
In this section we provide a brief and somewhat simplified overview of actual causality. The reader is referred to~\cite{HP01b,Hal16}
for the full exposition.

Actual causality~\cite{HP01b,Hal16} provides a formal framework for reasoning about the causes of observed outcomes. The framework describes the world in terms of variables and their values. 
A causal model $M$ consists of a set of variables $\V$ and a set of structural equations capturing the
dependencies between these variables. Some (salient) features of causal models can be represented by
a \emph{causal network}: a directed acyclic graph, with the nodes being the variables of $M$ and the directed
edges capturing the existence and direction of causal dependence. 


A \emph{context} $\vec{u}$ is an assignment to the variables of $M$ that do not have incoming edges in the
causal network (that is, their values do not depend on the values of other variables). $\vec{u}$ determines,
through the equations of $M$, the values of all variables in $\V$.
Together, a causal model $M$ and a context $\vec{u}$, which specifies the particular situation being analyzed, form a causal setting $(M,\vec{u})$.
We write $(M,\vec{u}) \sat \varphi$ if a Boolean formula $\varphi$ holds in the causal setting $(M,\vec{u})$.

Causal analysis is performed through \emph{interventions}. An intervention $\vec{X}\leftarrow\vec{x}$ assigns values $\vec{x}$ to a subset $\vec{X}\subseteq\V$, replacing their equations with these assignments. We write $(M,\vec{u})\models[\vec{X}\leftarrow\vec{x}]\varphi$ if $\varphi$ holds after this intervention. Interventions therefore allow us to reason about how changing selected variables affects an observed outcome.

We adapt the definitions from~\cite{icra} and restate them here as a basis for our extension in \Cref{sec:extending-ac}. Given a black-box CPS controller $g$ and a failure scenario $s$, we define a binary causal model $M_{g,s}$ as follows. The set $\V=\vec{V}\cup\{O\}$ of variables consists of one binary variable for each obstacle in $\env$ and a binary outcome variable $O$. Essentially, $\vec{V}$ is a Boolean \emph{mask} that indicates which obstacles are present and which are removed, allowing us to intervene on parts of $\env$. The outcome variable $O$ indicates whether the outcome of an execution under this masking matches that of the original scene. That is, $$O = [\status(\statetraj,\env)\?\status(\statetraj',\env')],$$
where $\env'$ is the environment induced by the (possibly intervened) causal setting, and $\statetraj'$ is its resulting trajectory. This results in a depth-$2$ causal model, where each obstacle is (causally) independent of the others.

Given the causal model $M_{g,s}$, let $\vec{u}_1$ be the context
that assigns $1$ to all variables in $\vec{V}$, and let $\vec{u}_0$ be the context that assigns $0$ to all these variables, where $1$ means the obstacle is present at its original location, and $0$ means the obstacle is removed. In other words, $\vec{u}_1$ matches the original crash scenario, and $\vec{u}_0$ matches the empty track. Following \cite{icra}, we define an explanation as follows.

\dfn[Sufficient Explanation~\cite{icra}]\label{defn:simple-exp}
A subset $\vec{V}_{\exp}$ of $\vec{V}$ is a \emph{sufficient explanation} of the outcome $O = 1$ 
if the following conditions hold:
\begin{description}
\item[{\rm EX1.}] $(M_{g,s},\vec{u}_0) \models [\vec{V}_{\exp} \leftarrow 1](O=1)$,
where $\vec{V}_{\exp} \leftarrow 1$ stands for assigning $1$ to all variables in 
$\vec{V}_{\exp}$.
\item[{\rm EX2.}] $\vec{V}_{\exp}$ is minimal; there is no strict subset $\vec{V}'_{exp}$ of $\vec{V}_{\exp}$ that satisfies EX1, where $\vec{v}'$ is the restriction of $\vec{v}$ to the variables in $\vec{V}'_{exp}$.
\end{description}
\edfn

Since each variable in $\vec{V}$ corresponds to an obstacle in scenario $s$, a sufficient explanation is one-to-one mapped to a subset of obstacles that are sufficient to reproduce the original failure. In other words, an explanation is a minimal subset of obstacles in a given scenario $s$ that is \emph{sufficient} for the failure.

Importantly, there can be more than one minimal explanation~\cite{CKK25}. The minimality condition in EX2 applies to each explanation locally (with respect to subset inclusion). Consequently, distinct subsets of obstacles may each be sufficient to reproduce the same failure, while no strict subset of either is sufficient.

The precise computation of explanations is intractable~\cite{CKKS26}. Consequently, existing explanation-generation approaches use heuristic search strategies~\cite{CKKS26,icra,kelly2026guess}. These strategies are based on the notion of the \emph{degree of responsibility}~\cite{CH04}, which quantifies causal influence. Formally, the degree of responsibility introduced in~\cite{CH04} quantifies the notion of an \emph{actual cause}. Thus, we first introduce the definition of an actual cause and then the degree of responsibility.

\dfn[Actual Cause \cite{Hal15}]\label{def:AC}
A subset $\vec{V}_{\ac} \subseteq \vec{V}$ and its values $\vec{V}_{\ac}$ is an 
\emph{actual cause} of $\varphi$ in
$(M_{g,s},\vec{u})$ if the following conditions hold:
\begin{description}
\item[{\rm AC1.}] $(M_{g,s},\vec{u}) \models (\vec{V}_{\ac} = \vec{V}_{\ac})\wedge\varphi$. In other words,
in the context $\vec{u}$, indeed $\vec{V}_{\ac} = \vec{V}_{\ac}$ and $\varphi$ holds.
\item[{\rm AC2.}] There exists an assignment $\vec{v}'_{ac}$ such that 
$(M_{g,s},\vec{u}) \models [\vec{V}_{\ac} \leftarrow \vec{v}'_{ac}]\neg{\varphi}$. That is, changing the values of
$\vec{V}_{\ac}$ to $\vec{v}'_{ac}$ alone is sufficient to change the value of the outcome, if
all other variables remain unchanged.
\item[{\rm AC3.}] $\vec{V}_{\ac}$ is minimal; there is no strict subset $\vec{V}'$ of $\vec{V}_{\ac}$ that satisfies AC2.
\end{description}
As in~\cite{Hal15}, we refer to parts of actual causes as causes.
\edfn

\dfn[Responsibility \cite{CH04}]\label{def:simple-resp}
The \emph{degree of responsibility} of $v_i$ for $\varphi$ is defined as
$1/|\vec{V}_{n}|$, where $\vec{V}_{\ac} \subseteq \vec{V}$ is a \emph{smallest} actual cause for $\varphi$ that 
contains $v_i$. 
If there is no actual cause containing $v_i$, its degree of responsibility is defined as $0$.
\edfn

We note that the degree of responsibility is defined with respect to the \emph{necessary} condition, that is, subsets of obstacles whose removal changes the outcome. This is in contrast to the sufficient explanation, which is a subset of obstacles that \emph{by itself} is sufficient for the outcome. The reason we choose the necessary condition for responsibility is mostly practical: the algorithm we later present approximates it from a set of simulations in which crashes are rare.


\section{Extending Actual Causality with Distance}
\label{sec:extending-ac}

In this section, we extend actual causality by incorporating a \emph{distance measure} into the framework. This extension is general and can be applied to any causal model, including the binary model described in \Cref{subsec:cause}. However, when restricted to binary variables, the use of a distance function becomes degenerate. Therefore, to properly account for the continuous characteristics of the examined CPS, we introduce a new causal model that reflects its underlying continuous structure.

A graphical representation of the updated CPS model is shown in \Cref{fig:causal_model}. Like the binary model described in \Cref{subsec:cause}, this is a depth-2 causal model with a similar overall structure. The key difference lies in the set of variables, $\vec{V}$, which now represent the obstacle locations, with separate variables for their $x, y$ coordinates.
To align with previous definitions, we explicitly add the valuation $\times$ for $x_i$ and $y_i$, to indicate the absence of the respective obstacle---i.e., ``$\times$'' corresponds to the value ``$0$'' in the binary model (\cref{defn:simple-exp}). 
As in the binary model, we denote a setting in which all obstacles are absent as $u_0$.
The output variable $O$ remains unchanged and indicates the outcome of executing the CPS controller under the given obstacle configuration.

The definition of sufficient explanation (\cref{defn:simple-exp}) carries over unchanged to the revised causal model. In particular, a sufficient explanation reproduces the original locations of obstacles in it, as it
explains the crash in terms of what actually occurred in the failure scenario.


We now present updated definitions of actual cause and responsibility by incorporating information about \emph{how far} an intervention deviates from the original scenario $\vec{u}_s$. Intuitively, rather than considering only the number of variables that cause an outcome, we can also account for the magnitude of the required change, as measured by a predefined distance metric. This allows us to prefer explanations that achieve the desired outcome with smaller deviations from the original state, thereby capturing a notion of proximity-based minimality. We first define a distance-weighted notion of actual cause, and then introduce a corresponding quantitative measure of a distance-weighted responsibility.

\dfn[Distance-Weighted Actual Cause]\label{defn:distance-weighted-ac}
Given some predefined distance metric $d \colon \vec{v} \times \vec{v} \to \mathbb{R}$ and scenario $\vec{u}_{s}$, a subset $\vec{V}_{ac} \subseteq \vec{V}$ and its values $\vec{v}_{ac}$ is a 
\emph{distance-weighted actual cause} of the outcome $\varphi$ in
$(M_{g,s},\vec{u}_s)$ if the following conditions hold:
\begin{description}
\item[{\rm DAC1.}] $(M_{g,s},\vec{u}_s) \models (\vec{V}_{ac} = \vec{v}_{ac})\wedge \varphi$. In other words,
in the context $\vec{u}_s$, indeed $\vec{V}_{ac} = \vec{v}_{ac}$ and $\varphi$ holds.
\item[{\rm DAC2.}] There exists an assignment $\vec{v}'_{ac}$ such that 
$(M_{g,s},\vec{u}_s) \models [\vec{V}_{ac} \leftarrow \vec{v}'_{ac}]\neg{\varphi}$. That is, changing the values of
$\vec{V}_{ac}$ to $\vec{v}'_{ac}$ alone is sufficient to change the value of the outcome, if
all other variables remain unchanged.
\item[{\rm DAC3.}] $\vec{V}_{ac}$ and $\vec{v}'_{ac}$ are minimal with respect to $d$; 
that is, there is no subset $\vec{V}'$ of $\vec{V}_{ac}$ that satisfies DAC2 with an assignment $\vec{v}''$ such that $d(\vec{v}'_{ac},\vec{v}_{s}) > d(\vec{v}'',\vec{v}_{s})$, where:
\begin{itemize}
    \item $\vec{v}_{s}$ is the assignment to $\vec{V}$ induced by $\vec{u}_{s}$,
    \item $\vec{v}'_{ac}$ is viewed as the assignment of $\vec{V}_{ac}$ to $\vec{v}'_{ac}$ and the rest of the variables in $\vec{V}$ according to $\vec{u}_{s}$, and
    \item $\vec{v}''$ is viewed as the assignment of $\vec{V}'$ to $\vec{v}''$ and the rest of the variables in $\vec{V}$ according to $\vec{u}_{s}$.
\end{itemize} 
\end{description}
\edfn

\dfn[Distance-Weighted Responsibility]\label{def:distance-weighted-resp}
Given some predefined distance metric $d \colon \vec{v} \times \vec{v} \to \mathbb{R}$, 
the \emph{distance-weighted degree of responsibility} of $v_i$ for the outcome $\varphi$ is defined as
$1/(1+d(\vec{v}_n,\vec{v}_{s}))$, where $\varphi$ is the original outcome, 
$\vec{V}_n \subseteq \vec{V}$ is the closest (according to $d$) 
distance-weighted actual cause for $\varphi$ that contains $v_i$,
$\vec{v}_n$ is the assignment of $\vec{V}_n$ that satisfies \Cref{defn:distance-weighted-ac},
and $\vec{v}_{s}$ is the assignment of $\vec{V}$ in the original scenario. 
If there is no distance-weighted actual cause that contains $v_i$, its degree of responsibility is defined as $0$.
\edfn

\begin{observation}
    \Cref{def:AC} and \Cref{def:simple-resp} are special cases of \Cref{defn:distance-weighted-ac} and \Cref{def:distance-weighted-resp}, respectively, obtained when the distance function simply counts the number of variables in $\vec{V}_{ac}$ (cardinality of the actual cause). Also, assuming non-negativity of the distance function, the new definition preserves the original responsibility value range of $[0,1]$, where $1$ represents the highest possible importance, and $0$ the lowest.
\end{observation}



\setlength{\fboxsep}{1pt}  
\begin{figure*}[ht]
    \begin{subfigure}{0.495\linewidth}
        \centering
        \fbox{\includegraphics[scale=0.22,trim=23pt 35pt 23pt 35pt,clip]{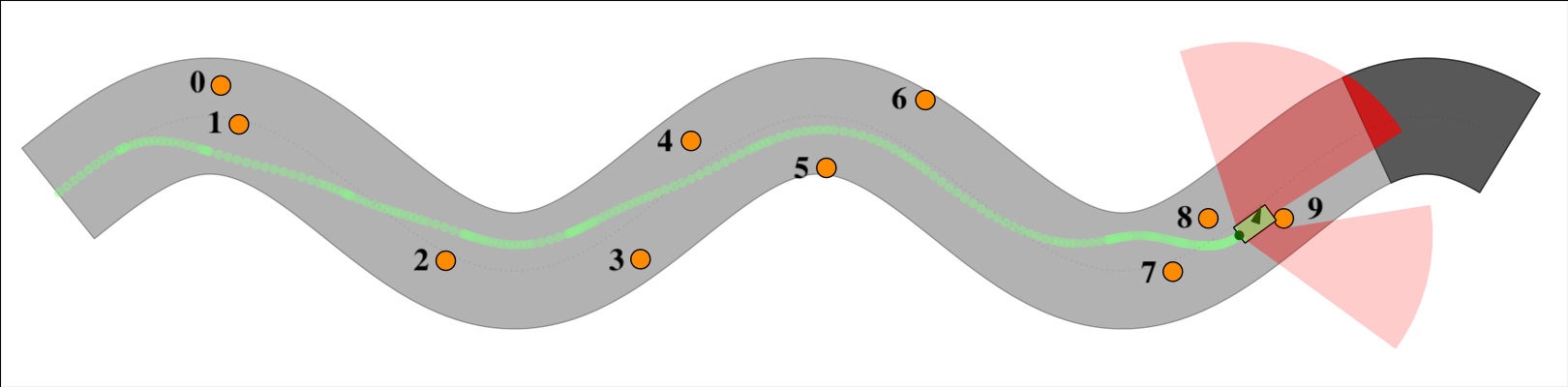}}
    \end{subfigure}
        \begin{subfigure}{0.495\linewidth}
        \centering
        \fbox{\includegraphics[scale=0.22,trim=23pt 35pt 23pt 35pt,clip]{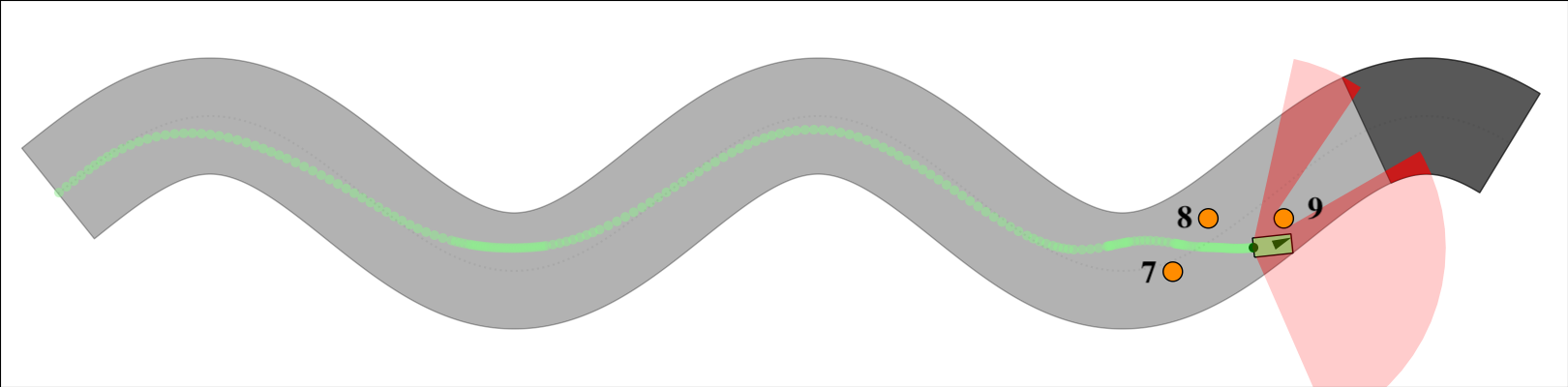}}
    \end{subfigure}
    \caption{Motivating example for distance-weighted responsibility: a failure scenario (left) and its sufficient explanation (right).}
        \label{fig:motivating-example}
\end{figure*}

\textbf{Motivating example:}
We illustrate the advantages of distance-weighted responsibility using the example shown in \Cref{fig:motivating-example}. It depicts a crash scene with a single explanation consisting of obstacles $7$, $8$, and $9$. The discrete formulation of responsibility considers only obstacle presence/absence. Consequently, several obstacles may receive the same responsibility value, resulting in a coarse ranking that provides limited guidance toward the obstacles in the explanation. In this example, obstacles $0$, $5$, $7$, $8$, and $9$ all have the responsibility $1$, reflecting
the fact that it is sufficient to remove any one of them to avoid the crash. Thus, they cannot be distinguished by the discrete ranking. In contrast, distance-weighted responsibility accounts for how much the obstacle configuration must change to alter the outcome, producing a more fine-grained ranking. Our distance-weighted approximation with Euclidean distance, based on $200$ samples, assigns responsibility values of $0.36$, $0.38$, $0.55$, $0.90$, and $0.66$ to obstacles $0$, $5$, $7$, $8$, and $9$, respectively, thereby prioritizing the three obstacles comprising an explanation.

\section{Responsibility-Guided Explanation-Derivation Algorithm}
\label{sec:method}

In this section, we present an algorithm for deriving sufficient explanations of a CPS failure, in the context of our running example, based on the extended definitions and causal model introduced in \Cref{sec:extending-ac}. We begin with a given crash scene $s$, for which we seek a sufficient explanation. In our setting, the causal model has depth two and is therefore relatively simple. Consequently, there is no need to explicitly construct it. Instead, the algorithm implicitly complies with the model through its search procedure.

As mentioned, the precise computation of explanations is intractable~\cite{CKKS26}. Thus we propose to derive explanations by ranking obstacles according to their \emph{approximate} degree of distance-weighted responsibility, prioritizing those with higher responsibility in the search process. This ranking serves as a heuristic that guides explanation discovery toward candidates that are more likely to be relevant. As a result, the algorithm is expected to reduce the number of simulation runs required to obtain an explanation. 

Previous work~\cite{icra} that applied actual causality to CPSs used a responsibility-guided algorithm over a binary causal model, using \Cref{def:simple-resp} and restricting interventions to obstacle removal as the only allowed type. This approximation is quite coarse and may result in rankings that are unhelpful for guiding the explanation search (see motivating example in~\cref{fig:motivating-example}). Here we propose a more fine-grained approach that approximates the distance-weighted responsibility over the continuous causal model we introduce in \cref{sec:extending-ac}. Later, in \cref{sec:evaluation}, we demonstrate that this approach indeed reduces the overall effort required to compute explanations.


The responsibility approximation algorithm is presented in \Cref{alg:responsibility}. It creates a dataset of $m$ random interventions over the original crash scene $s$ and considers them in increasing order of distance (lines~\ref{line:create-interventions}--\ref{line:sort-interventions}). Following \Cref{defn:distance-weighted-ac}, interventions that change the original outcome (DAC2) and satisfy the minimality condition (DAC3) are identified as causes. The responsibility of their elements is then updated according to \Cref{def:distance-weighted-resp} in the loop in line~\ref{line:loop-resp}.

\begin{algorithm}[t]
\small
\caption{Distance-Weighted Responsibility Approximation}
\label{alg:responsibility}
\renewcommand{\algorithmicrequire}{\textbf{Input:}}
\begin{algorithmic}[1]
    \Require{Crash scene $s=(\env,\state_0)$, element set \emph{elements}, a distance function $d$, samples budget $m$}
    \State{resp$(e) \gets 0$ for every $e$ in \emph{elements}}
    \State{causes $\gets \emptyset$}
    \State{$\statetraj \gets \text{run\_simulation($s$)}$}
    \State{$E \gets \text{elements\_observed}(\statetraj, \env)$}
    \State{\label{line:create-interventions}$\mathcal{I} \gets \text{sample\_m\_random\_interventions}(E, m)$}
    \State{\label{line:sort-interventions}$\mathcal{I} \gets \text{sort\_by\_distance}(\mathcal{I}, d)$}
    \For{$s'=(\env',\state_0)$ in $\mathcal{I}$}
        \State{$\statetraj' \gets \text{run\_simulation($s'$)}$}
        \If{$\status(\statetraj,\env) \ne \status(\statetraj',\env')$} \Comment{DAC2 (\cref{defn:distance-weighted-ac})}
            \State{$S \gets \text{get\_intervened\_elements}(\env, \env')$}
            \If{$\nexists S' \in \text{causes} \text{ such that } S' \subseteq S$} \Comment{DAC3 (\cref{defn:distance-weighted-ac})}
                \State{causes $\gets$ causes $\cup S$}
                \For{\label{line:loop-resp}$e$ in $S$}\Comment{Responsibility (as in \cref{def:distance-weighted-resp})}
                    \State{resp$(e) \gets \text{max}(\text{resp}(e), 1/(1+d(\env,\env')))$}
                \EndFor
            \EndIf
        \EndIf
    \EndFor
    \State{\Return{resp}}
\end{algorithmic}
\end{algorithm}

The distance-weighted responsibility-guided explanation
search algorithm is presented in \Cref{alg:explanation}. It ranks obstacles according to
their approximate responsibility computed using
\Cref{alg:responsibility} and
incrementally evaluates prefixes of this ordering until the original
failure is reproduced. This subset serves as a candidate explanation.
The algorithm then searches its powerset for a smaller explanation (lines~\ref{line-min-loop-starts}--\ref{line-min-loop-ends}).

\begin{algorithm}[t]
\small
\caption{Distance-weighted responsibility-guided explanation search algorithm}
\label{alg:explanation}
\renewcommand{\algorithmicrequire}{\textbf{Input:}}
\begin{algorithmic}[1]
        \Require{Crash scene $s=(\env,\state_0)$, responsibility distance function $d$, samples budget $m$}
    \State{$elements \gets \text{element\_set\_in\_the\_environment\_of\_example($\env$)}$}
    \State{$\statetraj \gets \text{run\_simulation($s$)}$}
    \Statex{\Comment{Responsibility-guided ranking of elements (\Cref{alg:responsibility}})}
    \State{$resp \gets\Call{Responsibility\_approximation}{elements, d, m}$}
    \State{$el\_sorted \gets \text{sort\_by\_responsibility}(elements,resp)$}
    \For{\label{line:exp-loop}$i$ in $\{ 1, \dots , |el\_sorted| \}$}\Comment{Expl. candidates evaluation}
        \State{$S \gets el\_sorted[0:i] $}\Comment{First $i$ ranked obstacles}
        \State{$s' \gets \text{intervene on $s$ by retaining only obstacles in $S$}$}
        \State{$\statetraj' \gets \text{run\_simulation($s'$)}$}
        \If{$\status(\statetraj,\env) \? \status(\statetraj',\env')$}
            \State{$V_{\exp} \gets \{ e \in E \mid e \text{ in } el\_sorted[:i] \}$}
            \For{\label{line-min-loop-starts}$S \in 2^{V_{\exp}}$}\Comment{Explanation minimization}
                \State{$s' \gets \text{intervene on $s$ by retaining only obstacles in $S$}$}
                \State{$\statetraj' \gets \text{run\_simulation($s'$)}$}
                \If{$\status(\statetraj,\env) = \status(\statetraj',\env')$}   
                    \If{$|S| < |V_{\exp}|$}   
                        \State{\label{line-min-loop-ends}$V_{\exp} \gets S$}
                    \EndIf
                \EndIf
            \EndFor
            \State{\textbf{break}}
        \EndIf
    \EndFor
    \State{\Return{$V_{\exp}$}}
\end{algorithmic}
\end{algorithm}

\cref{alg:explanation} is guaranteed to return a valid explanation, regardless of the quality of the responsibility approximation. The responsibility ranking is used only as a heuristic to determine the order in which obstacles are considered. In the worst case, the incremental search eventually considers all obstacles, thereby reproducing the original outcome. The subsequent minimization step then removes redundant obstacles until a minimal outcome-preserving subset remains, thereby satisfying both conditions of \cref{defn:simple-exp}. Thus, the responsibility approximation affects the efficiency of the search but not the validity of the resulting explanation. In the following section, we empirically demonstrate that the proposed distance-weighted responsibility improves the efficiency of finding both single and multiple explanations.

\section{Experimental Evaluation}
\label{sec:evaluation}

In this section, we quantitatively evaluate our approach to generating explanations for CPS failures. To evaluate the usefulness of our extensions to distance-weighted actual cause and responsibility, we use them to guide the explanation search, which serves as a proxy for measuring the quality of our approach. Since the algorithm is guaranteed to find a valid explanation, our evaluation focuses on efficiency, particularly the computational effort required to find one or multiple explanations when the search is guided by a distance-weighted responsibility ranking.

\begin{figure}[t]
\centering
\includegraphics[width=\columnwidth,height=3.8cm,trim=50pt 0pt 10pt 180pt,clip]{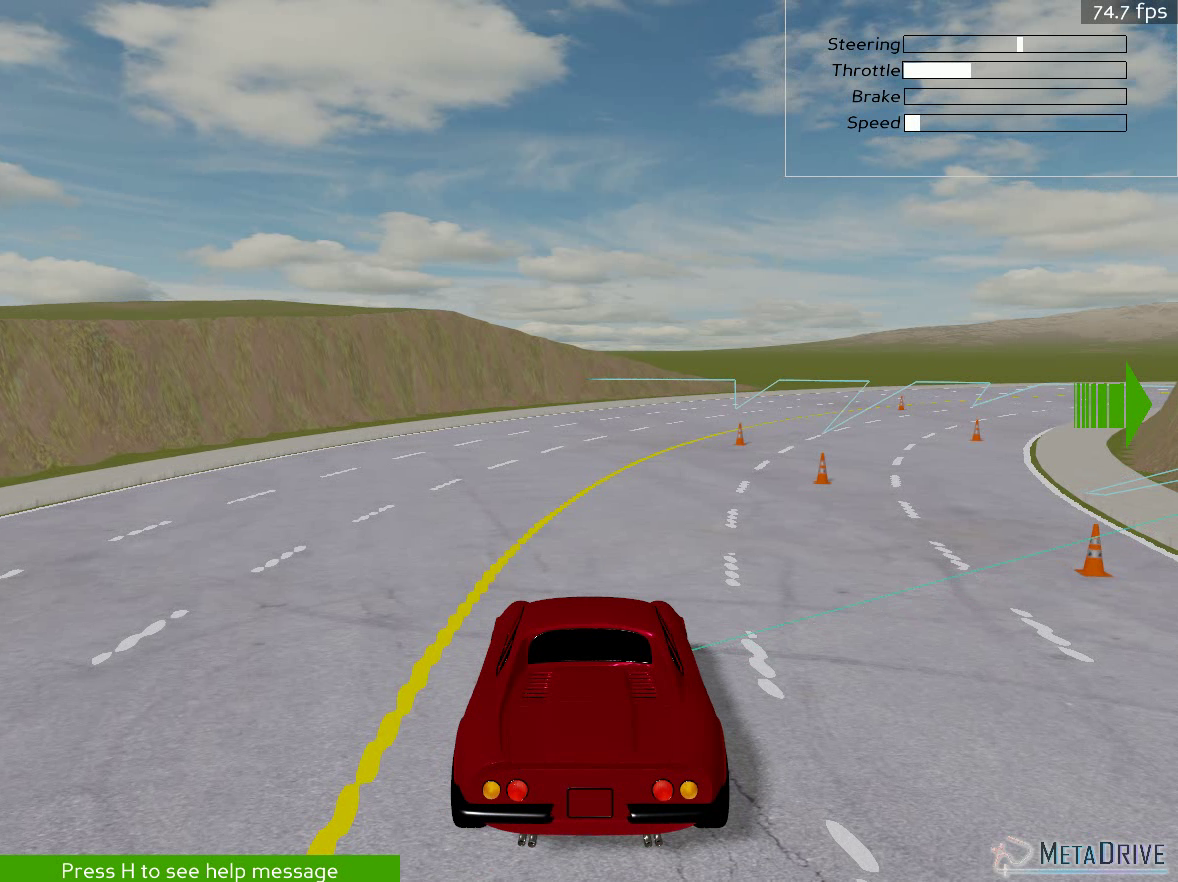}
\caption{The MetaDrive simulation~\cite{li2022metadrive}. The vehicle is tasked with navigating an obstructed track.}
\label{fig:metadrive}
\end{figure}

\noindent\textbf{Experimental Setup.} We evaluated our approach using two autonomous-driving simulators: LiteRacer~\cite{Elimelech24LiteRacer}, an open-source Python engine for simulating an autonomous vehicle on an obstructed track (see \cref{fig:car}), and MetaDrive~\cite{li2022metadrive}, an open-source driving simulator for AI and autonomy research (see \cref{fig:metadrive}). We used each simulator's default neural network-based vehicle controller. To evaluate generalizability across environments, we considered several track geometries and configurations. In LiteRacer, we defined three geometries: (1) \textbf{Sinusoidal}, following a sine-wave shape; (2) \textbf{Segmented}, combining straight and curved segments; and (3) \textbf{Lane Shift}, consisting primarily of straight segments connected by turns resembling lane shifts. We evaluated each geometry under four configurations combining short/long tracks and narrow/wide widths. In MetaDrive, we defined three geometries: (1) \textbf{Curve}, (2) \textbf{Straight}, and (3) \textbf{Mixed}, combining straight and curved segments. Visualizations of all evaluated tracks are provided at~\cite{supplementarymaterial}.

For each track type, we created $100$ failing examples. To assess how our method performs as environment complexity grows, we increased the number of obstacles from 10 to 14, with 20 examples for each level. To ensure that additional obstacles made the explanation task more complex, we focused solely on examples where the vehicle observed all obstacles during execution. This ensured that each obstacle could affect the vehicle's behavior and potentially be included in an explanation. Experiments were conducted on an Intel(R) Xeon(R) Platinum 8468 with 64GB RAM.

\pgfplotsset{
    exhaustive/.style={
        color=black!70,
        very thick,
        mark=*,
        mark size=1.3pt
    },
    discrete50/.style={
        color=red!50,
        thick,
        mark=square*,
        mark size=1.3pt
    },
    discrete100/.style={
        color=red!100,
        thick,
        mark=square*,
        mark size=1.3pt
    },
    euclidean50/.style={
        color=blue!50,
        thick,
        mark=triangle*,
        mark size=1.5pt
    },
    euclidean100/.style={
        color=blue!100,
        thick,
        mark=triangle*,
        mark size=1.5pt
    }
}

\newcommand{\plotruns}[1]{%
    \addplot[exhaustive]
        table[x=obstacles,y=exhaustive,col sep=comma] {#1};

    \addplot[discrete50]
        table[x=obstacles,y=discrete_50,col sep=comma] {#1};

    \addplot[discrete100]
        table[x=obstacles,y=discrete_100,col sep=comma] {#1};

    \addplot[euclidean50]
        table[x=obstacles,y=euclidean_50,col sep=comma] {#1};

    \addplot[euclidean100]
        table[x=obstacles,y=euclidean_100,col sep=comma] {#1};
}

\newcommand{\plotrunswithlegend}[1]{%
    \addplot[exhaustive]
        table[x=obstacles,y=exhaustive,col sep=comma] {#1};
    \addlegendentry{Exhaustive}

    \addplot[discrete50]
        table[x=obstacles,y=discrete_50,col sep=comma] {#1};
    \addlegendentry{Discrete 50}

    \addplot[discrete100]
        table[x=obstacles,y=discrete_100,col sep=comma] {#1};
    \addlegendentry{Discrete 100}

    \addplot[euclidean50]
        table[x=obstacles,y=euclidean_50,col sep=comma] {#1};
    \addlegendentry{Euclidean 50}

    \addplot[euclidean100]
        table[x=obstacles,y=euclidean_100,col sep=comma] {#1};
    \addlegendentry{Euclidean 100}
}

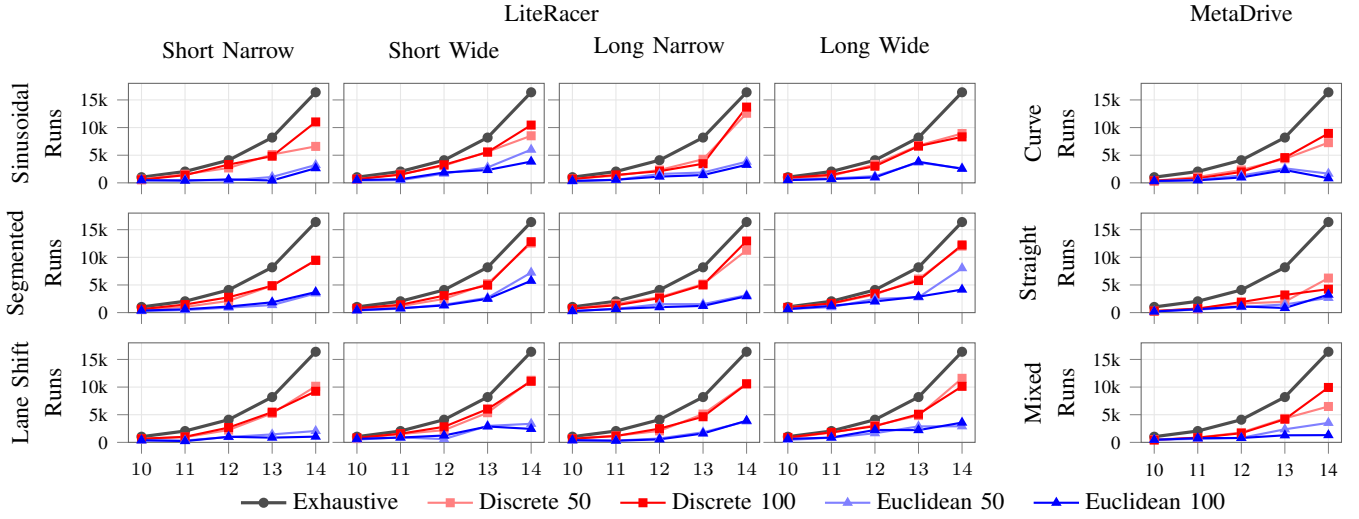
\begin{figure*}[t]
    \centering

    \begin{tikzpicture}

        \begin{groupplot}[
            group style={
                group name=plots,
                group size=6 by 3,
                horizontal sep=2mm,
                vertical sep=4mm,
            },
            width=0.238\textwidth,
            height=0.163\textwidth,
            xmin=9.7,
            xmax=14.3,
            ymin=0,
            ymax=18000,
            xtick={10,11,12,13,14},
            ytick={0,5000,10000,15000},
            yticklabels={0,5k,10k,15k},
            scaled y ticks=false,
            tick align=outside,
            tick pos=left,
            grid=major,
            grid style={black!10},
            axis line style={black!60},
            tick label style={font=\scriptsize},
            label style={font=\small},
            title style={font=\small},
        ]


        \nextgroupplot[
            title={Short Narrow},
            ylabel={Runs},
            xticklabels=\empty,
            legend columns=5,
            legend to name=commonlegend,
            legend style={
                font=\small,
                draw=none,
                /tikz/every even column/.append style={
                    column sep=3mm
                }
            }
        ]
        \plotrunswithlegend{results/sinusoidal_short_narrow_runs_results.csv}

        \nextgroupplot[
            title={Short Wide},
            xticklabels=\empty,
            yticklabels=\empty
        ]
        \plotruns{results/sinusoidal_short_wide_runs_results.csv}

        \nextgroupplot[
            title={Long Narrow},
            xticklabels=\empty,
            yticklabels=\empty
        ]
        \plotruns{results/sinusoidal_long_narrow_runs_results.csv}

        \nextgroupplot[
            title={Long Wide},
            xticklabels=\empty,
            yticklabels=\empty
        ]
        \plotruns{results/sinusoidal_long_wide_runs_results.csv}

        \nextgroupplot[
            hide axis,
            width=0.19\textwidth
        ]

        \nextgroupplot[
            ylabel={Runs},
            xticklabels=\empty
        ]
        \plotruns{results/curve_runs_results.csv}


        \nextgroupplot[
            ylabel={Runs},
            xticklabels=\empty
        ]
        \plotruns{results/segmented_short_narrow_runs_results.csv}

        \nextgroupplot[
            xticklabels=\empty,
            yticklabels=\empty
        ]
        \plotruns{results/segmented_short_wide_runs_results.csv}

        \nextgroupplot[
            xticklabels=\empty,
            yticklabels=\empty
        ]
        \plotruns{results/segmented_long_narrow_runs_results.csv}

        \nextgroupplot[
            xticklabels=\empty,
            yticklabels=\empty
        ]
        \plotruns{results/segmented_long_wide_runs_results.csv}

        \nextgroupplot[
            hide axis,
            width=0.19\textwidth
        ]

        \nextgroupplot[
            ylabel={Runs},
            xticklabels=\empty
        ]
        \plotruns{results/straight_runs_results.csv}


        \nextgroupplot[
            ylabel={Runs}
        ]
        \plotruns{results/lane_shift_short_narrow_runs_results.csv}

        \nextgroupplot[
            yticklabels=\empty
        ]
        \plotruns{results/lane_shift_short_wide_runs_results.csv}

        \nextgroupplot[
            yticklabels=\empty
        ]
        \plotruns{results/lane_shift_long_narrow_runs_results.csv}

        \nextgroupplot[
            yticklabels=\empty
        ]
        \plotruns{results/lane_shift_long_wide_runs_results.csv}

        \nextgroupplot[
            hide axis,
            width=0.19\textwidth
        ]

        \nextgroupplot[
            ylabel={Runs},
            xtick={10,11,12,13,14},
            xticklabels={10,11,12,13,14}
        ]
        \plotruns{results/mixed_runs_results.csv}

        \end{groupplot}


        \node[
            anchor=south,
            font=\small
        ] at ([yshift=7mm]
             $(plots c1r1.north)!0.5!(plots c4r1.north)$)
        {LiteRacer};

        \node[
            anchor=south,
            font=\small
        ] at ([yshift=7mm]
             $(plots c6r1.north)$)
        {MetaDrive};


        \node[
            rotate=90,
            anchor=south,
            font=\small
        ] at ([xshift=-12mm]$(plots c1r1.west)$)
        {Sinusoidal};

        \node[
            rotate=90,
            anchor=south,
            font=\small
        ] at ([xshift=-12mm]$(plots c1r2.west)$)
        {Segmented};

        \node[
            rotate=90,
            anchor=south,
            font=\small
        ] at ([xshift=-12mm]$(plots c1r3.west)$)
        {Lane Shift};


        \node[
            rotate=90,
            anchor=south,
            font=\small
        ] at ([xshift=-12mm]$(plots c6r1.west)$)
        {Curve};

        \node[
            rotate=90,
            anchor=south,
            font=\small
        ] at ([xshift=-12mm]$(plots c6r2.west)$)
        {Straight};

        \node[
            rotate=90,
            anchor=south,
            font=\small
        ] at ([xshift=-12mm]$(plots c6r3.west)$)
        {Mixed};


        \node[
            anchor=north
        ] at ([yshift=-4mm]
             $(plots c1r3.south)!0.5!(plots c6r3.south)$)
        {\pgfplotslegendfromname{commonlegend}};

    \end{tikzpicture}

    \caption{Efficiency of explanation generation algorithms as a function of the number of obstacles ($10$--$14$), measured by the number of simulation runs (lower is better).}
    \label{fig:runs-comparison}

\end{figure*}

\noindent\textbf{Implementation Details.} For the intervention sampling, we implemented a process in which, for each sample, each obstacle is selected for intervention independently with some predefined probability. Each selected obstacle is perturbed by adding standard normal Gaussian noise centered at its original location. Full details can be found at~\cite{supplementarymaterial}. To demonstrate generality, we evaluated the approach using two distance functions. The first is the standard Euclidean distance between obstacle locations in the two simulations. The second is the Chamfer distance~\cite{Fan_2017_CVPR}, which measures the distance between two point sets by averaging the distance from each point in one set to its nearest point in the other set, symmetrized over both directions. These metrics capture different similarity notions: one based on correspondence between individual obstacles and the other on the geometry of the obstacle sets, illustrating that the framework can accommodate user-defined distance functions based on the characteristics of the CPS.

\noindent\textbf{Comparison.} To assess our contribution, we compared our approach with the discrete responsibility mechanism proposed in prior work~\cite{icra}, which uses a binary causal model similar to the one outlined in \Cref{subsec:cause}. In this model, each obstacle is represented by a binary variable indicating its presence or absence; thus, interventions are restricted to obstacle removal, and responsibility is defined according to \Cref{def:simple-resp}. For both approaches, we approximated responsibility using sample budgets of $50$ and $100$. As a baseline, we also report the simulation runs required by the exhaustive approach discussed in~\cite{icra}, which exhaustively searches all subsets. The complete code for the evaluations and the implementation of \Cref{alg:explanation} are publicly available at~\cite{supplementarymaterial}.

To measure the computational effort required in each method, we calculated the overall number of simulation runs it required, as this constitutes the dominant factor that determines the computational cost in practice. All other components (e.g., distance calculations, algorithm logic) are negligible in comparison and do not materially affect the overall runtime. We empirically validate this observation in an additional analysis provided at~\cite{supplementarymaterial}.

\noindent\textbf{Results.} \Cref{fig:runs-comparison} reports the mean number of simulation runs required to find an explanation using the exhaustive approach, the discrete responsibility ranking, and our distance-weighted responsibility ranking with the Euclidean distance. Due to space limitations and to keep the presentation focused, we report the corresponding results for the Chamfer distance at~\cite{supplementarymaterial}, where the advantage of the distance-weighted approach is largely preserved. Across the different environments, the responsibility-guided approaches generally require fewer simulation runs than the exhaustive search, with the difference becoming more pronounced as the number of obstacles increases. More importantly, the distance-weighted ranking consistently outperforms the discrete ranking, often substantially, demonstrating that accounting for the magnitude of interventions provides more effective guidance for the explanation search. Increasing the sampling budget from 50 to 100 further improves the distance-weighted approach in most settings, with the additional cost of responsibility estimation remaining small relative to the resulting reduction in search effort.

\noindent\textbf{Multiple explanation evaluation.} We further evaluated the quality of the responsibility ranking across multiple explanations. For each failure scenario, we computed all explanations and measured their coverage as obstacles are incrementally added in order of their responsibility ranking. At each point in the ranking, coverage is defined as the fraction of explanations fully contained in the set of obstacles considered so far. We summarize this behavior using the area under the coverage curve (AUC), with higher values indicating a more effective ranking.

\Cref{fig:responsibility-coverage} and \Cref{tab:auc-results} compare the discrete and Euclidean distance-weighted responsibility rankings in prioritizing obstacles across multiple explanations. \Cref{fig:responsibility-coverage} visualizes the ranking curves for selected settings. The distance-weighted ranking generally achieves higher explanation coverage earlier in the ranking, indicating that obstacles relevant to multiple explanations tend to receive higher priority. \Cref{tab:auc-results} summarizes the results of all evaluated settings using AUC. The distance-weighted ranking achieves a higher AUC in almost all settings, often by a substantial margin, further demonstrating that incorporating distance provides more effective guidance not only for finding a single explanation but also for identifying multiple explanations. This advantage is consistent across the evaluated numbers of obstacles and is particularly pronounced in the LiteRacer environments. Results using the Chamfer distance, and visualization of the ranking curves for all settings, are provided at~\cite{supplementarymaterial} and show the same overall advantage of the distance-weighted ranking.

\begin{table}[t]
    \scriptsize
    \centering

    \caption{%
        AUC of explanation coverage for the discrete and distance-weighted responsibility rankings (higher is better).
    }
    \label{tab:auc-results}

    \scriptsize
    \setlength{\tabcolsep}{2.5pt}

    \begin{tabular}{c c || cc cc cc cc cc}

        \toprule
        \multicolumn{2}{c}{} &
        \multicolumn{2}{c}{\textbf{10}} &
        \multicolumn{2}{c}{\textbf{11}} &
        \multicolumn{2}{c}{\textbf{12}} &
        \multicolumn{2}{c}{\textbf{13}} &
        \multicolumn{2}{c}{\textbf{14}}
        \\
        \cmidrule(lr){3-4}\cmidrule(lr){5-6}\cmidrule(lr){7-8}\cmidrule(lr){9-10}\cmidrule(lr){11-12}
        & \textbf{Env.} &
        \textbf{Disc.} & \textbf{Dist.} &
        \textbf{Disc.} & \textbf{Dist.} &
        \textbf{Disc.} & \textbf{Dist.} &
        \textbf{Disc.} & \textbf{Dist.} &
        \textbf{Disc.} & \textbf{Dist.}
        \\
        \midrule


        \multirow{12}{*}{\rotatebox[origin=c]{90}{LiteRacer}}
        & Sin-SN
        & 0.21 & \textbf{0.29}
        & 0.15 & \textbf{0.35}
        & 0.10 & \textbf{0.45}
        & 0.19 & \textbf{0.40}
        & 0.14 & \textbf{0.30}
        \\

        & Sin-SW
        & 0.19 & \textbf{0.35}
        & 0.17 & \textbf{0.32}
        & 0.12 & \textbf{0.29}
        & 0.12 & \textbf{0.24}
        & 0.13 & \textbf{0.31}
        \\

        & Sin-LN
        & 0.21 & \textbf{0.36}
        & 0.17 & \textbf{0.31}
        & 0.20 & \textbf{0.35}
        & 0.18 & \textbf{0.32}
        & 0.08 & \textbf{0.30}
        \\

        & Sin-LW
        & 0.13 & \textbf{0.35}
        & 0.15 & \textbf{0.35}
        & 0.14 & \textbf{0.32}
        & 0.10 & \textbf{0.23}
        & 0.17 & \textbf{0.33}
        \\
        \cmidrule(lr){2-12}

        & Seg-SN
        & 0.19 & \textbf{0.40}
        & 0.20 & \textbf{0.33}
        & 0.13 & \textbf{0.42}
        & 0.19 & \textbf{0.40}
        & 0.17 & \textbf{0.38}
        \\

        & Seg-SW
        & 0.21 & \textbf{0.34}
        & 0.18 & \textbf{0.34}
        & 0.15 & \textbf{0.29}
        & 0.14 & \textbf{0.28}
        & 0.11 & \textbf{0.29}
        \\

        & Seg-LN
        & 0.22 & \textbf{0.43}
        & 0.20 & \textbf{0.34}
        & 0.19 & \textbf{0.38}
        & 0.12 & \textbf{0.39}
        & 0.11 & \textbf{0.34}
        \\

        & Seg-LW
        & 0.13 & \textbf{0.28}
        & 0.16 & \textbf{0.29}
        & 0.11 & \textbf{0.26}
        & 0.12 & \textbf{0.24}
        & 0.09 & \textbf{0.25}
        \\
        \cmidrule(lr){2-12}

        & LS-SN
        & 0.23 & \textbf{0.40}
        & 0.19 & \textbf{0.40}
        & 0.14 & \textbf{0.46}
        & 0.12 & \textbf{0.28}
        & 0.17 & \textbf{0.37}
        \\

        & LS-SW
        & 0.18 & \textbf{0.20}
        & 0.11 & \textbf{0.19}
        & 0.13 & \textbf{0.34}
        & 0.11 & \textbf{0.29}
        & 0.12 & \textbf{0.26}
        \\

        & LS-LN
        & 0.19 & \textbf{0.37}
        & 0.17 & \textbf{0.33}
        & 0.17 & \textbf{0.40}
        & 0.20 & \textbf{0.38}
        & 0.20 & \textbf{0.39}
        \\

        & LS-LW
        & 0.15 & \textbf{0.31}
        & 0.12 & \textbf{0.31}
        & 0.15 & \textbf{0.28}
        & 0.13 & \textbf{0.33}
        & 0.16 & \textbf{0.39}
        \\
        \cmidrule(lr){2-12}


        \multirow{3}{*}{\rotatebox[origin=c]{90}{MetaDr.}}
        & Curve
        & 0.34 & \textbf{0.38}
        & 0.28 & \textbf{0.46}
        & 0.20 & \textbf{0.39}
        & 0.15 & \textbf{0.33}
        & 0.21 & \textbf{0.44}
        \\

        & Straight
        & \textbf{0.41} & \textbf{0.41}
        & 0.28 & \textbf{0.33}
        & 0.22 & \textbf{0.33}
        & 0.20 & \textbf{0.36}
        & 0.25 & \textbf{0.30}
        \\

        & Mixed
        & 0.32 & \textbf{0.33}
        & 0.25 & \textbf{0.35}
        & 0.20 & \textbf{0.41}
        & 0.16 & \textbf{0.36}
        & 0.22 & \textbf{0.33}
        \\
        \bottomrule
    \end{tabular}

\end{table}

\pgfplotsset{
    discrete/.style={
        color=red!80,
        thick,
        mark=square*,
        mark size=1.2pt
    },
    distanceplot/.style={
        color=blue!80,
        thick,
        mark=triangle*,
        mark size=1.4pt
    }
}

\newcommand{\plotresponsibility}[1]{%
    \addplot[discrete]
        table[
            x=obstacles,
            y=discrete,
            col sep=comma
        ] {#1};

    \addplot[distanceplot]
        table[
            x=obstacles,
            y=distance,
            col sep=comma
        ] {#1};
}

\newcommand{\plotresponsibilitywithlegend}[1]{%
    \addplot[discrete]
        table[
            x=obstacles,
            y=discrete,
            col sep=comma
        ] {#1};
    \addlegendentry{Discrete}

    \addplot[distanceplot]
        table[
            x=obstacles,
            y=distance,
            col sep=comma
        ] {#1};
    \addlegendentry{Distance}
}

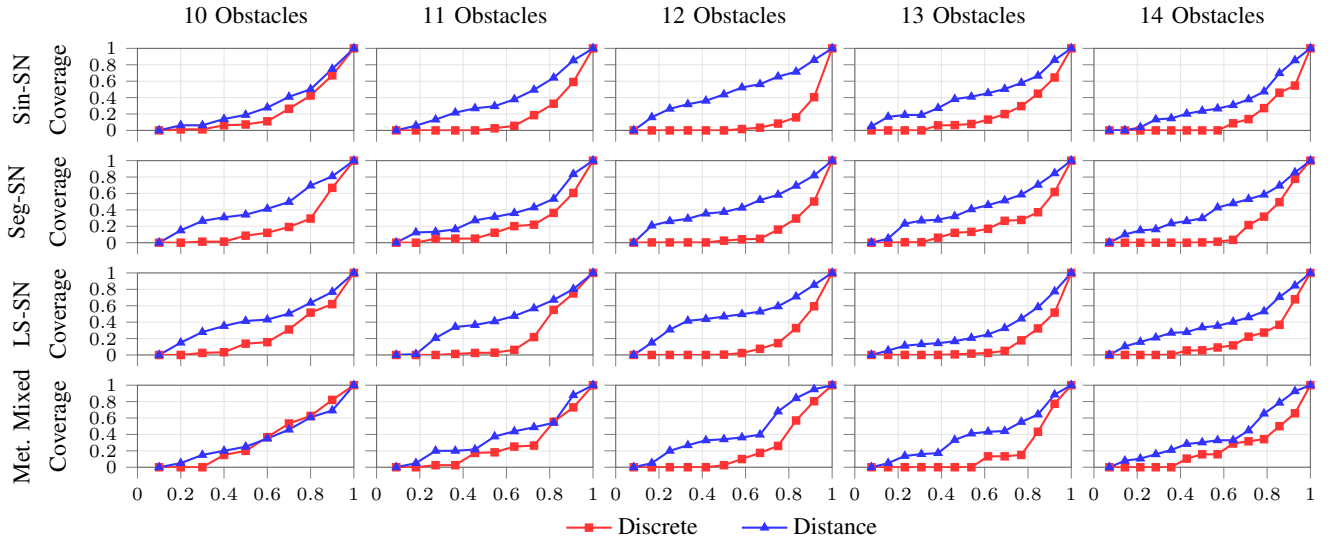
\begin{figure*}[t]
    \centering

    \begin{tikzpicture}

        \begin{groupplot}[
            group style={
                group name=respplots,
                group size=5 by 4,
                horizontal sep=3mm,
                vertical sep=4mm
            },
            width=0.25\textwidth,
            height=0.15\textwidth,
            xmin=0,
            xmax=1,
            ymin=0,
            ymax=1,
            xtick={0,0.2,0.4,0.6,0.8,1.0},
            ytick={0,0.2,0.4,0.6,0.8,1.0},
            tick align=outside,
            tick pos=left,
            grid=major,
            grid style={black!10},
            axis line style={black!60},
            tick label style={
                font=\scriptsize
            },
            label style={
                font=\small
            },
            title style={
                font=\small
            }
        ]


        \nextgroupplot[
            title={10 Obstacles},
            ylabel={Coverage},
            xticklabels=\empty,
            legend columns=2,
            legend to name=responsibilitylegend,
            legend style={
                font=\small,
                draw=none,
                /tikz/every even column/.append style={
                    column sep=5mm
                }
            }
        ]
        \plotresponsibilitywithlegend{results/sinusoidal_short_narrow_combined_responsibility_multiple_explanations_results_10.csv}

        \nextgroupplot[
            title={11 Obstacles},
            xticklabels=\empty,
            yticklabels=\empty
        ]
        \plotresponsibility{results/sinusoidal_short_narrow_combined_responsibility_multiple_explanations_results_11.csv}

        \nextgroupplot[
            title={12 Obstacles},
            xticklabels=\empty,
            yticklabels=\empty
        ]
        \plotresponsibility{results/sinusoidal_short_narrow_combined_responsibility_multiple_explanations_results_12.csv}

        \nextgroupplot[
            title={13 Obstacles},
            xticklabels=\empty,
            yticklabels=\empty
        ]
        \plotresponsibility{results/sinusoidal_short_narrow_combined_responsibility_multiple_explanations_results_13.csv}

        \nextgroupplot[
            title={14 Obstacles},
            xticklabels=\empty,
            yticklabels=\empty
        ]
        \plotresponsibility{results/sinusoidal_short_narrow_combined_responsibility_multiple_explanations_results_14.csv}


        \nextgroupplot[
            ylabel={Coverage},
            xticklabels=\empty
        ]
        \plotresponsibility{results/segmented_short_narrow_combined_responsibility_multiple_explanations_results_10.csv}

        \nextgroupplot[
            xticklabels=\empty,
            yticklabels=\empty
        ]
        \plotresponsibility{results/segmented_short_narrow_combined_responsibility_multiple_explanations_results_11.csv}

        \nextgroupplot[
            xticklabels=\empty,
            yticklabels=\empty
        ]
        \plotresponsibility{results/segmented_short_narrow_combined_responsibility_multiple_explanations_results_12.csv}

        \nextgroupplot[
            xticklabels=\empty,
            yticklabels=\empty
        ]
        \plotresponsibility{results/segmented_short_narrow_combined_responsibility_multiple_explanations_results_13.csv}

        \nextgroupplot[
            xticklabels=\empty,
            yticklabels=\empty
        ]
        \plotresponsibility{results/segmented_short_narrow_combined_responsibility_multiple_explanations_results_14.csv}


        \nextgroupplot[
            ylabel={Coverage},
            xticklabels=\empty
        ]
        \plotresponsibility{results/lane_shift_short_narrow_combined_responsibility_multiple_explanations_results_10.csv}

        \nextgroupplot[
            xticklabels=\empty,
            yticklabels=\empty
        ]
        \plotresponsibility{results/lane_shift_short_narrow_combined_responsibility_multiple_explanations_results_11.csv}

        \nextgroupplot[
            xticklabels=\empty,
            yticklabels=\empty
        ]
        \plotresponsibility{results/lane_shift_short_narrow_combined_responsibility_multiple_explanations_results_12.csv}

        \nextgroupplot[
            xticklabels=\empty,
            yticklabels=\empty
        ]
        \plotresponsibility{results/lane_shift_short_narrow_combined_responsibility_multiple_explanations_results_13.csv}

        \nextgroupplot[
            xticklabels=\empty,
            yticklabels=\empty
        ]
        \plotresponsibility{results/lane_shift_short_narrow_combined_responsibility_multiple_explanations_results_14.csv}


        \nextgroupplot[
            ylabel={Coverage}
        ]
        \plotresponsibility{results/metadrive_mixed_combined_responsibility_multiple_explanations_results_10.csv}

        \nextgroupplot[
            yticklabels=\empty
        ]
        \plotresponsibility{results/metadrive_mixed_combined_responsibility_multiple_explanations_results_11.csv}

        \nextgroupplot[
            yticklabels=\empty
        ]
        \plotresponsibility{results/metadrive_mixed_combined_responsibility_multiple_explanations_results_12.csv}

        \nextgroupplot[
            yticklabels=\empty
        ]
        \plotresponsibility{results/metadrive_mixed_combined_responsibility_multiple_explanations_results_13.csv}

        \nextgroupplot[
            yticklabels=\empty
        ]
        \plotresponsibility{results/metadrive_mixed_combined_responsibility_multiple_explanations_results_14.csv}

        \end{groupplot}




        \node[
            rotate=90,
            anchor=south,
            font=\small
        ] at ([xshift=-13mm]$(respplots c1r1.west)$)
        {Sin-SN};

        \node[
            rotate=90,
            anchor=south,
            font=\small
        ] at ([xshift=-13mm]$(respplots c1r2.west)$)
        {Seg-SN};

        \node[
            rotate=90,
            anchor=south,
            font=\small
        ] at ([xshift=-13mm]$(respplots c1r3.west)$)
        {LS-SN};

        \node[
            rotate=90,
            anchor=south,
            font=\small
        ] at ([xshift=-13mm]$(respplots c1r4.west)$)
        {Met. Mixed};


        \node[
            anchor=north
        ] at ([yshift=-4mm]
             $(respplots c1r4.south)!0.5!(respplots c5r4.south)$)
        {\pgfplotslegendfromname{responsibilitylegend}};

    \end{tikzpicture}

    \caption{%
        Multiple explanations coverage as a function of the fraction of obstacles considered, ordered by the discrete and distance-weighted responsibility rankings (higher is better).
    }
    \label{fig:responsibility-coverage}

\end{figure*}

\section{Conclusion}
\label{sec:conclusion}

We extended the theoretical framework of actual causality with the notion of distance-weighted actual cause and distance-weighted responsibility, enabling it to account for continuous inputs. While motivated by the continuous nature of CPSs, these definitions constitute a novel extension of the theory of actual causality in general, allowing causal influence to account for the magnitude of interventions rather than only their cardinality. We further introduced a system-agnostic (black-box) explanation algorithm that leverages distance-weighted responsibility to guide the explanation search. Our experimental evaluation across environments from two autonomous-driving simulators demonstrated that the distance-weighted responsibility provides more effective guidance than discrete responsibility, substantially reducing the computational effort required to find explanations.

\bibliographystyle{IEEEtran}
\bibliography{references,all}


\clearpage

\begin{appendices}
\section{Evaluated Tracks Visualization}
\label{app:tracks}

\begin{figure}[h]
\centering
\includegraphics[width=\columnwidth]{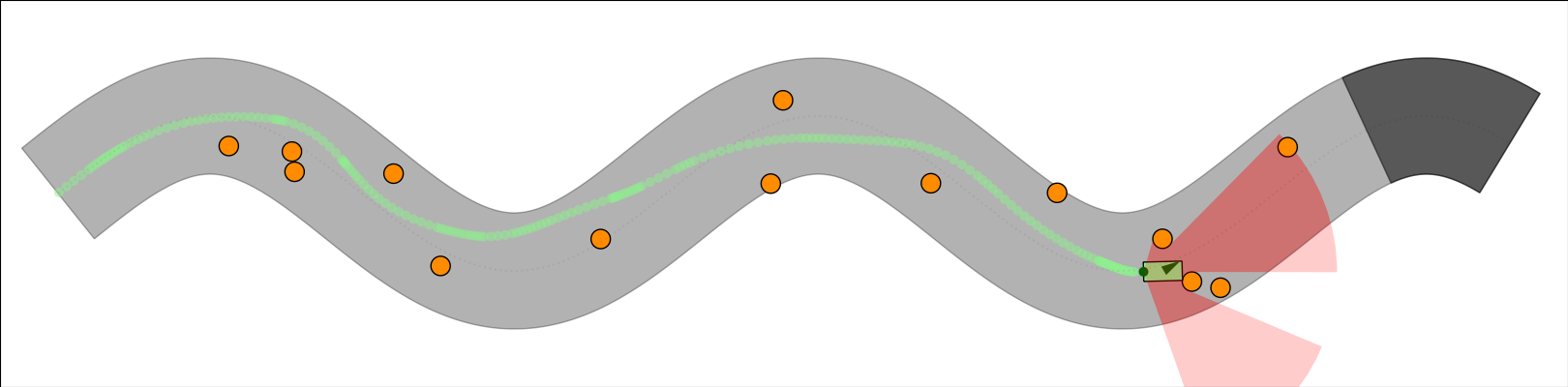}
\caption{Sinusoidal short narrow}
\label{fig:Sin-SN}
\end{figure}

\begin{figure}[h]
\centering
\includegraphics[width=\columnwidth]{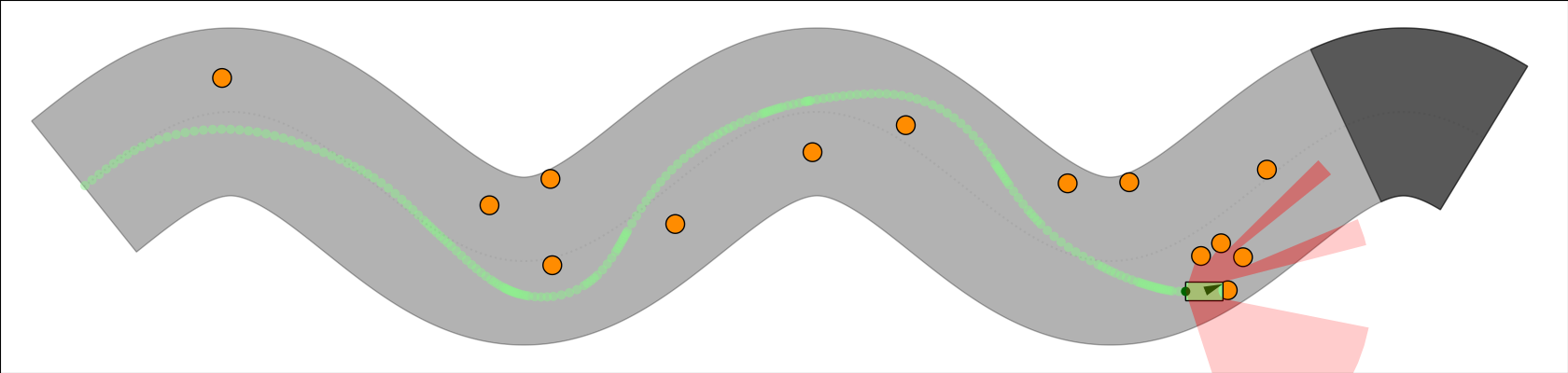}
\caption{Sinusoidal short wide}
\label{fig:Sin-SW}
\end{figure}

\begin{figure}[h]
\centering
\includegraphics[width=\columnwidth]{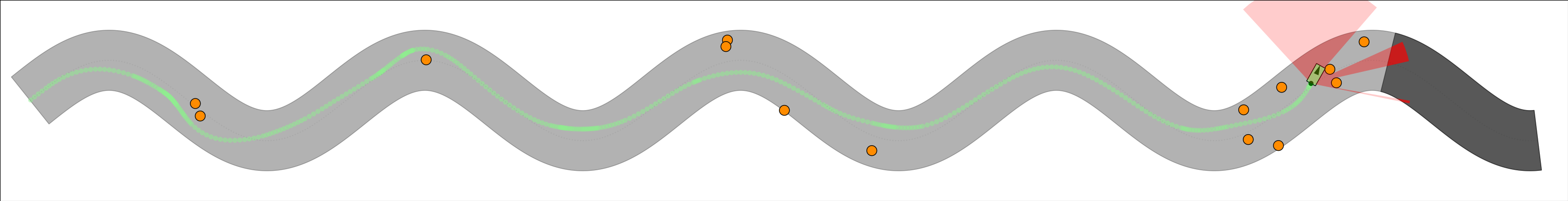}
\caption{Sinusoidal long narrow}
\label{fig:Sin-LN}
\end{figure}

\begin{figure}[h]
\centering
\includegraphics[width=\columnwidth]{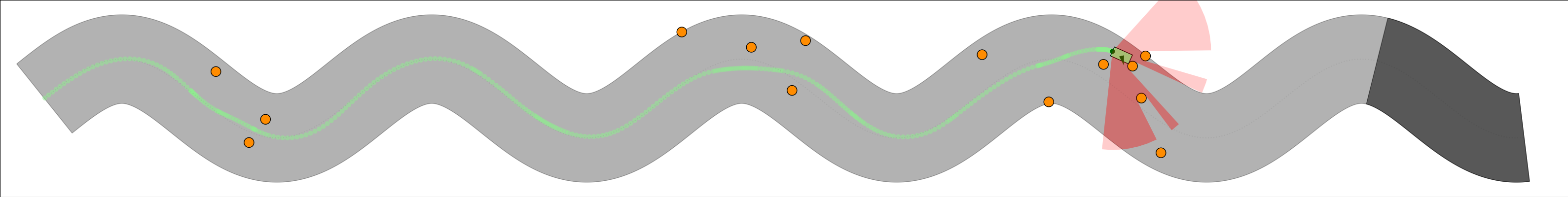}
\caption{Sinusoidal long wide}
\label{fig:Sin-LW}
\end{figure}

\begin{figure}[h]
\centering
\includegraphics[width=\columnwidth]{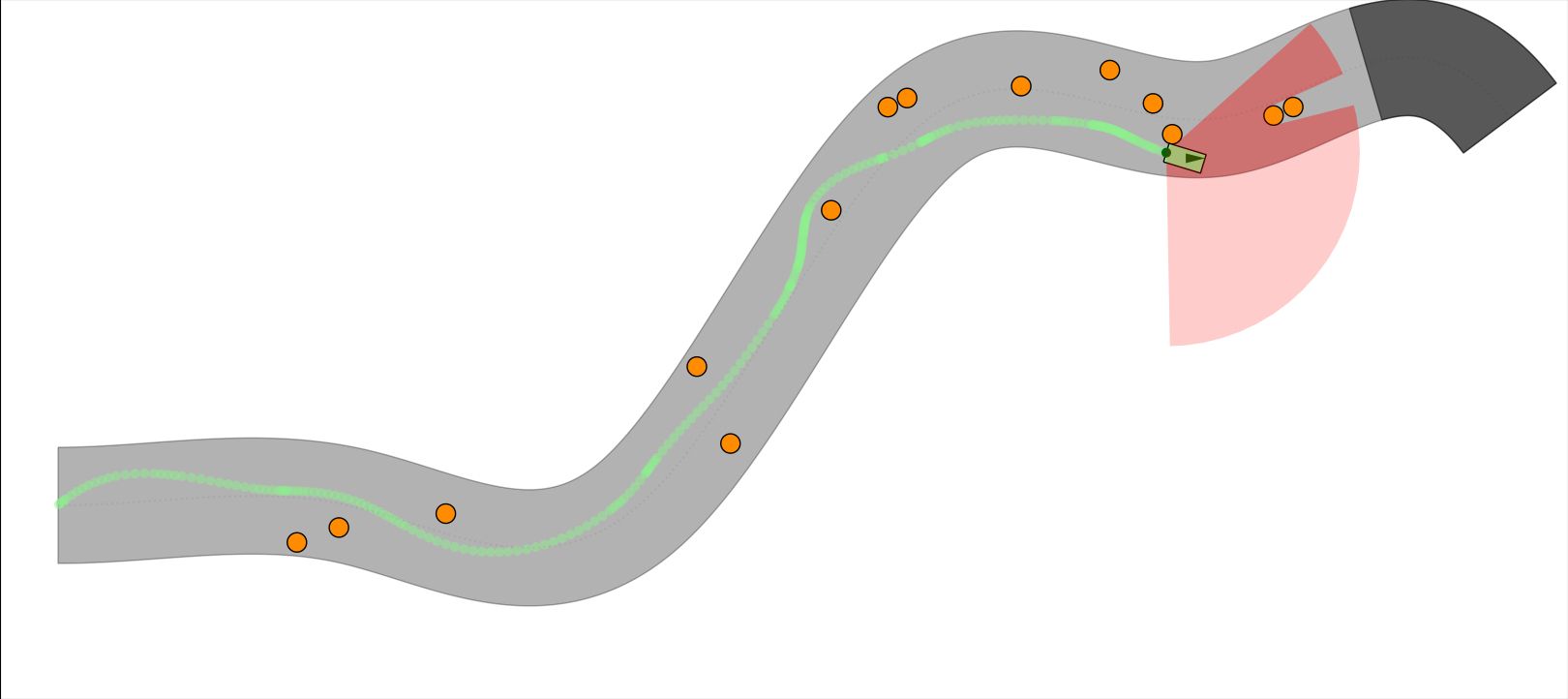}
\caption{Segmented short narrow}
\label{fig:Seg-SN}
\end{figure}

\begin{figure}[h]
\centering
\includegraphics[width=\columnwidth]{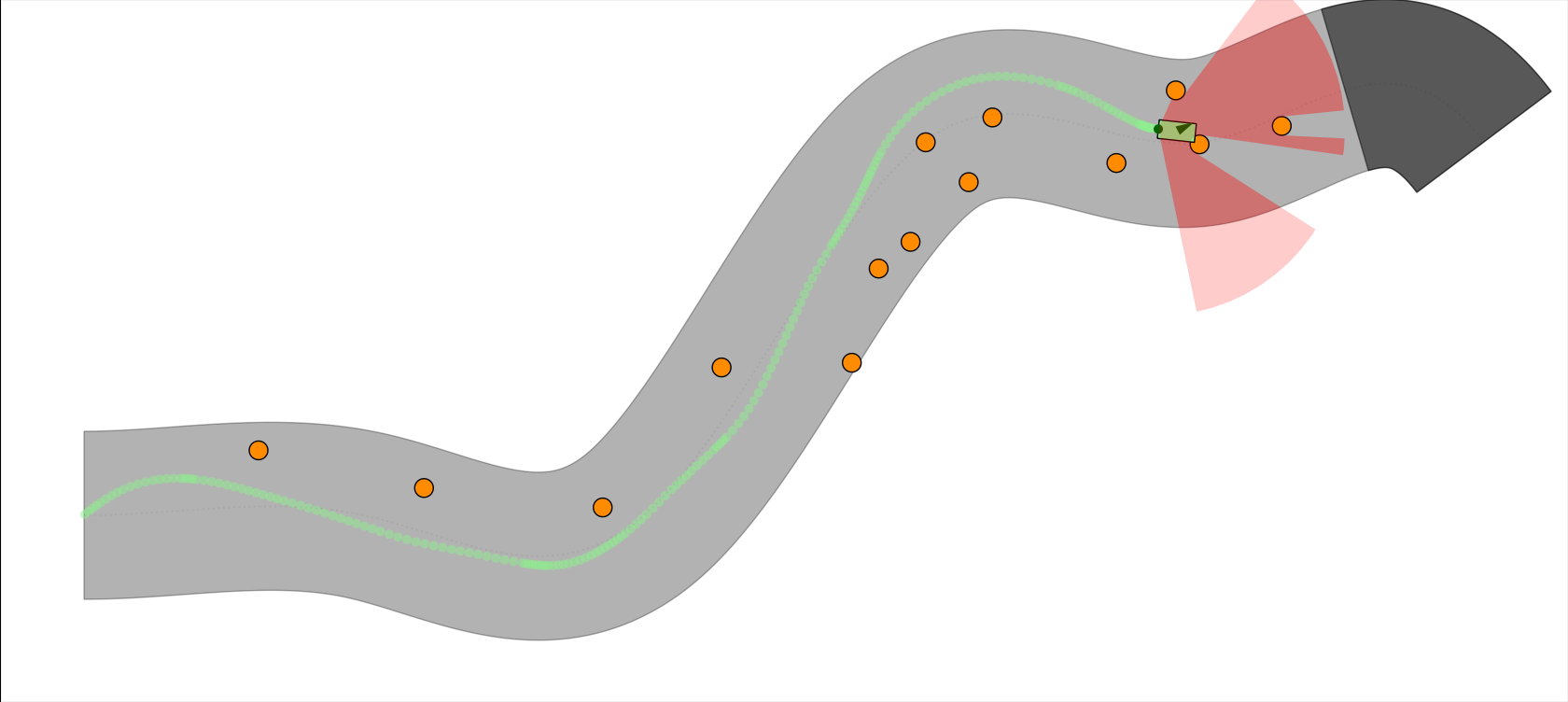}
\caption{Segmented short wide}
\label{fig:Seg-SW}
\end{figure}

\begin{figure}[h]
\centering
\includegraphics[width=\columnwidth]{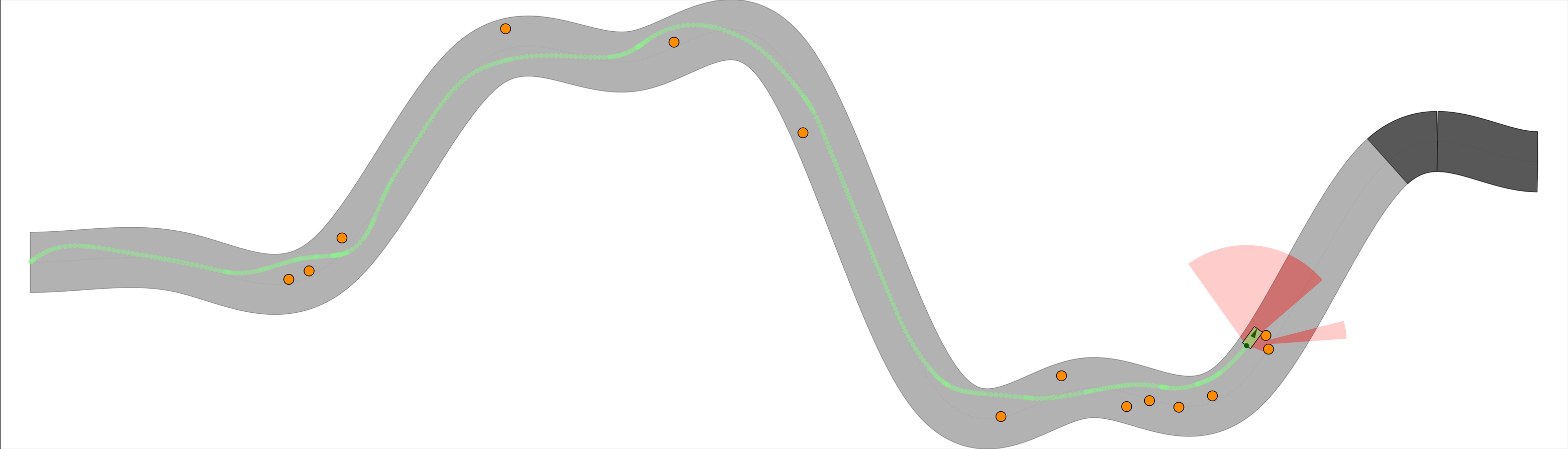}
\caption{Segmented long narrow}
\label{fig:Seg-LN}
\end{figure}

\begin{figure}[h]
\centering
\includegraphics[width=\columnwidth]{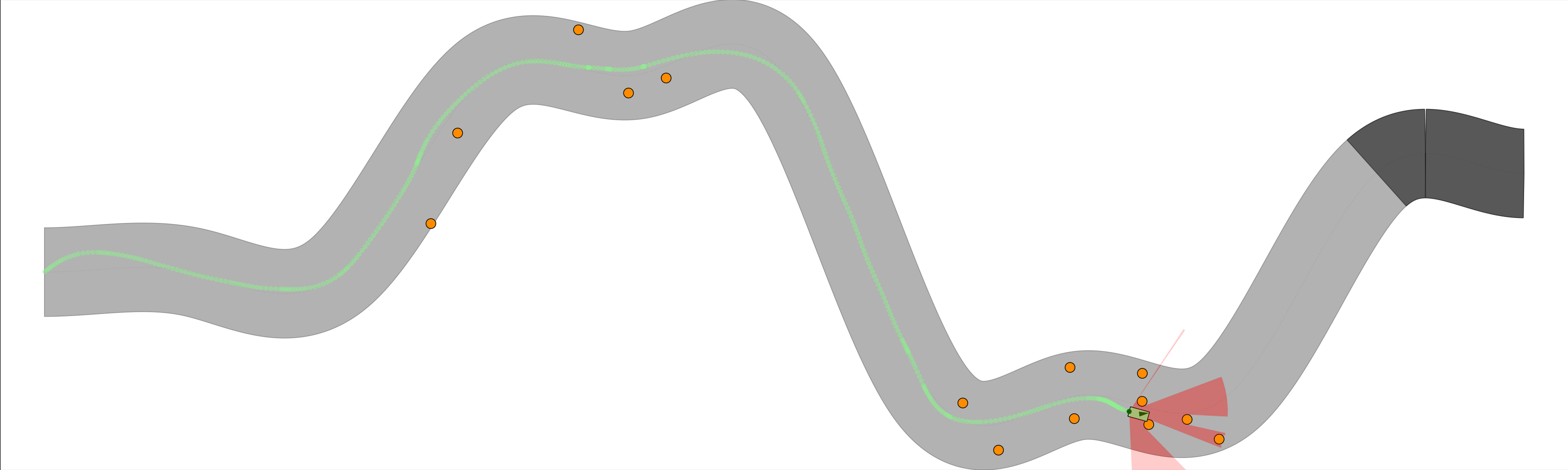}
\caption{Segmented long wide}
\label{fig:Seg-LW}
\end{figure}

\begin{figure}[h]
\centering
\includegraphics[width=\columnwidth]{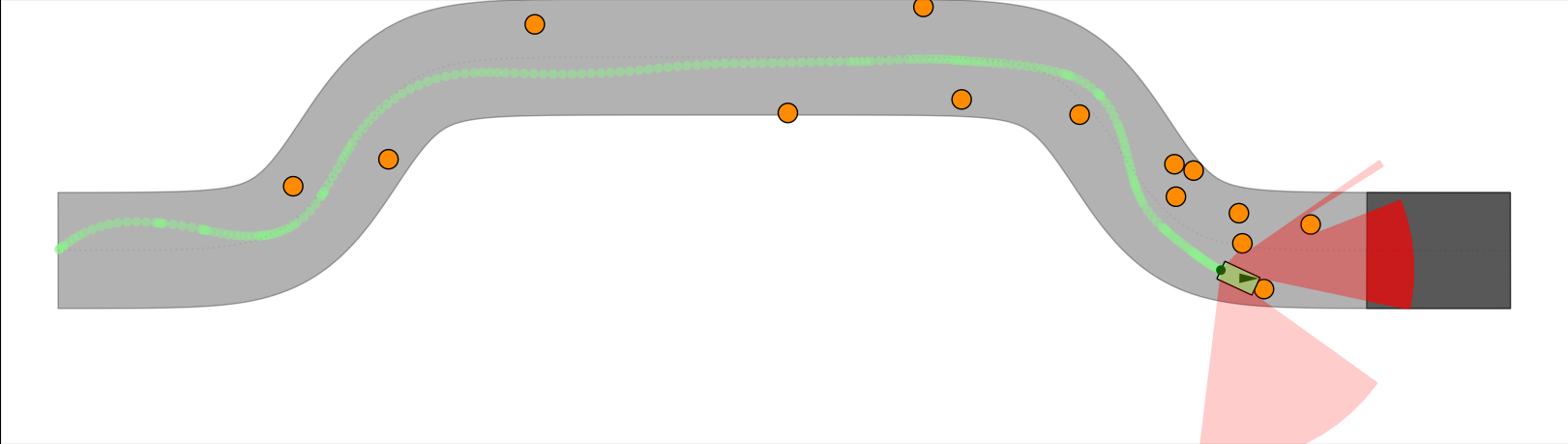}
\caption{Lane Shift short narrow}
\label{fig:LS-SN}
\end{figure}

\begin{figure}[h]
\centering
\includegraphics[width=\columnwidth]{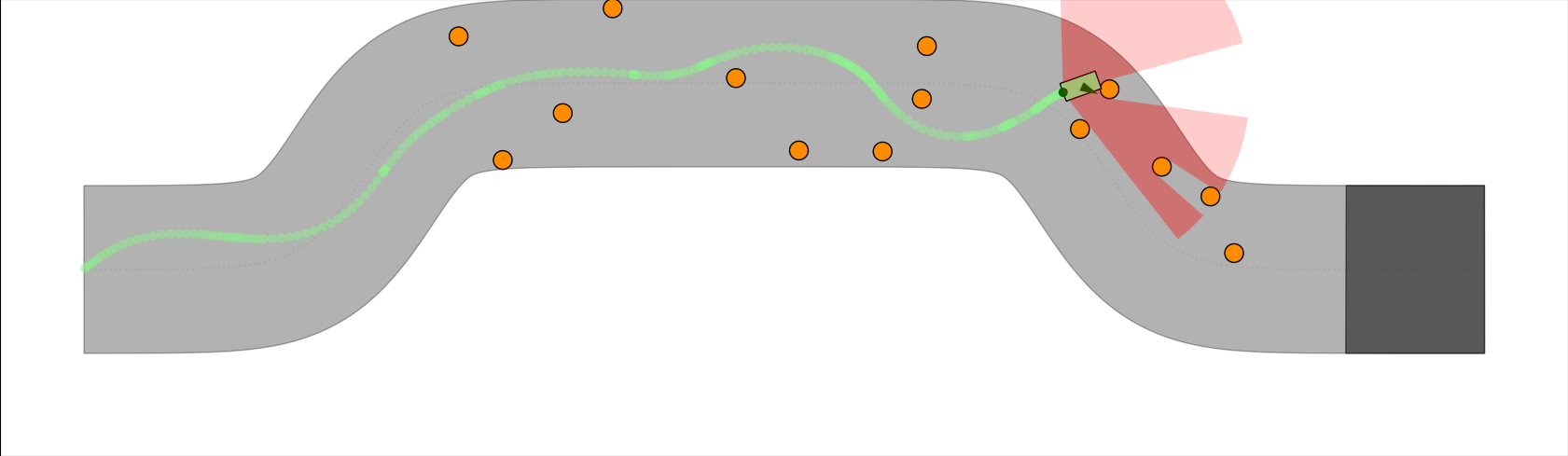}
\caption{Lane Shift short wide}
\label{fig:LS-SW}
\end{figure}

\begin{figure}[h]
\centering
\includegraphics[width=\columnwidth]{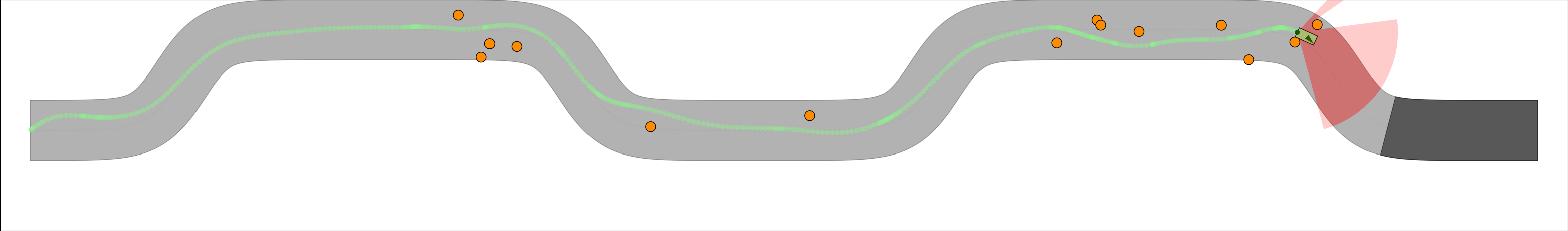}
\caption{Lane Shift long narrow}
\label{fig:LS-LN}
\end{figure}

\begin{figure}[h]
\centering
\includegraphics[width=\columnwidth]{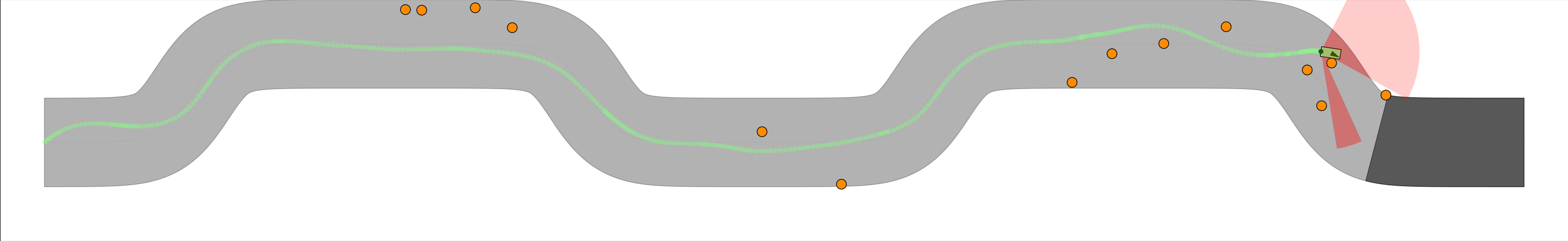}
\caption{Lane Shift long wide}
\label{fig:LS-LW}
\end{figure}

\begin{figure}[h]
\centering
\includegraphics[width=\columnwidth]{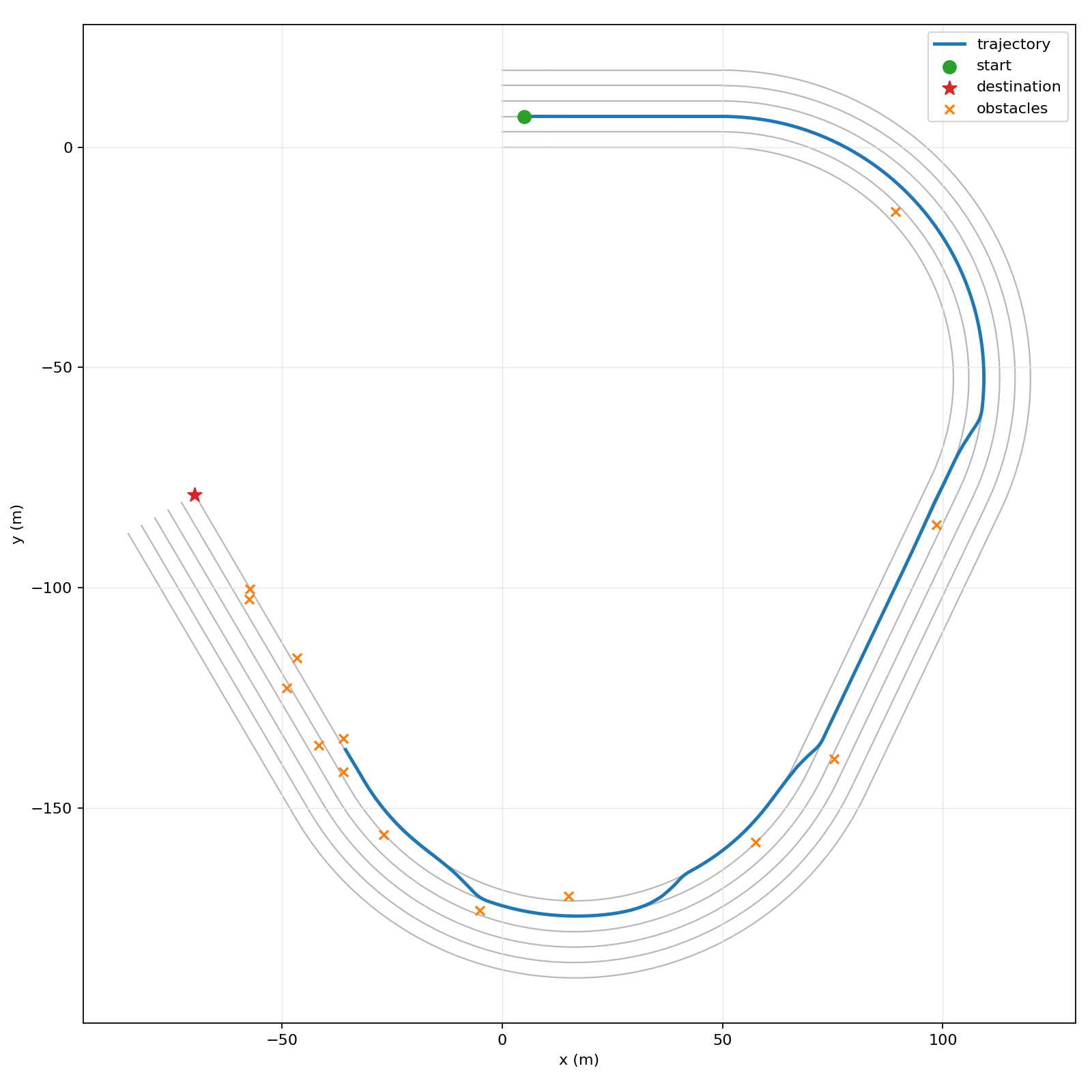}
\caption{MetaDrive Curve}
\label{fig:Curve}
\end{figure}

\begin{figure}[h]
\centering
\includegraphics[width=\columnwidth]{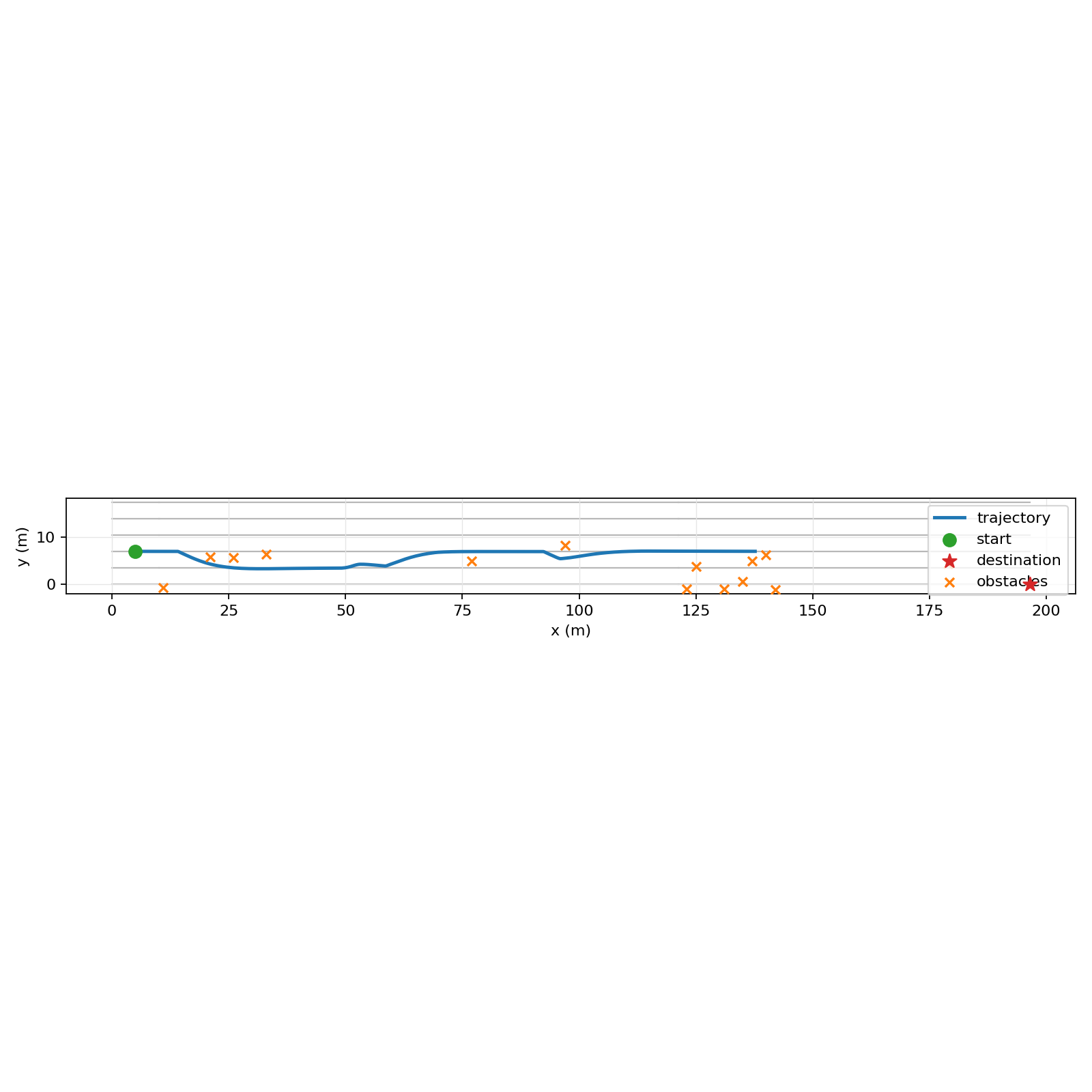}
\caption{MetaDrive Straight}
\label{fig:Straight}
\end{figure}

\begin{figure}[h]
\centering
\includegraphics[width=\columnwidth]{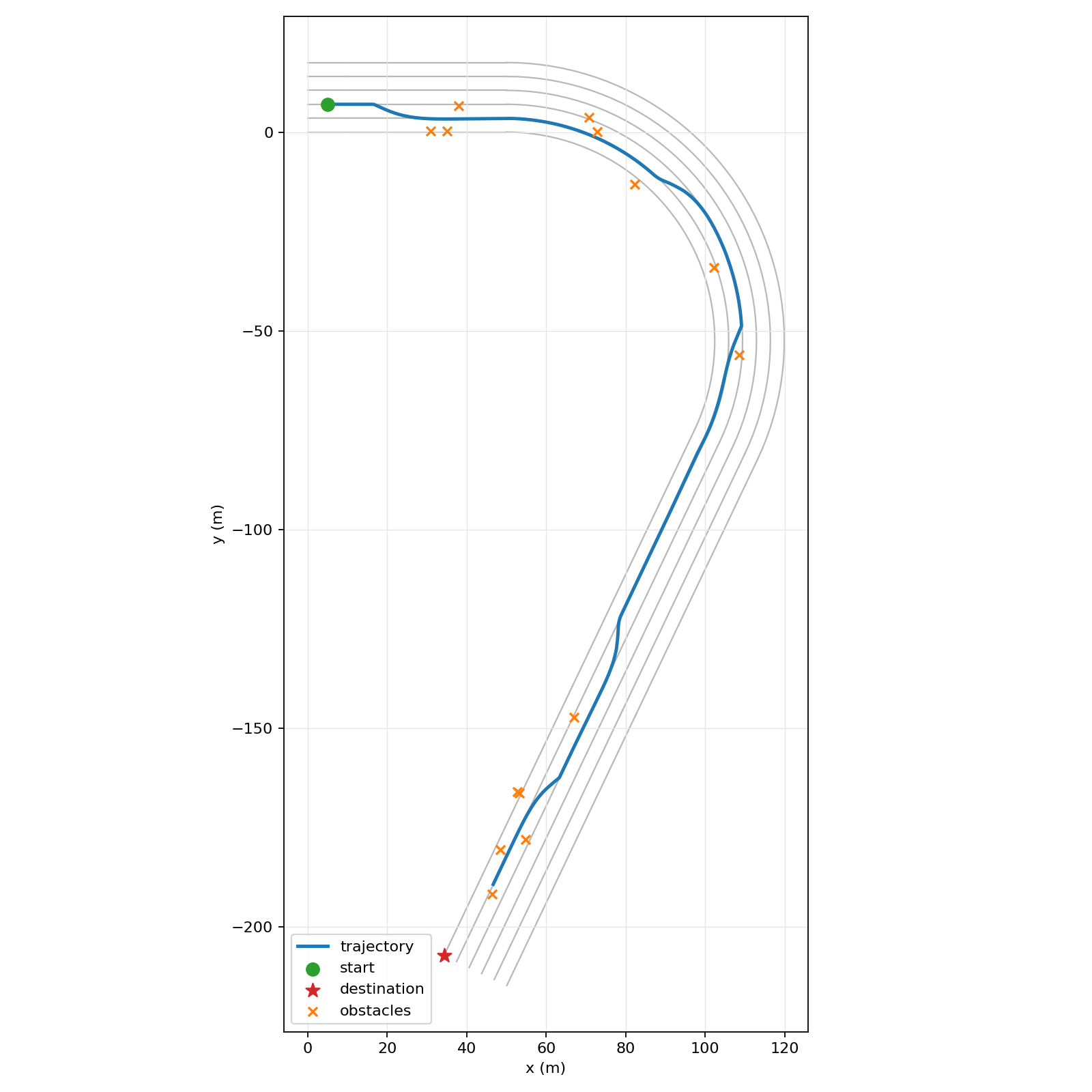}
\caption{MetaDrive Mixed}
\label{fig:Mixed}
\end{figure}

\crefrange{fig:Sin-SN}{fig:Mixed} show all types of evaluated tracks. We note that the visualizations of the MetaDrive environments (\crefrange{fig:Curve}{fig:Mixed}) \textbf{include lanes in both directions}. Hence, the vehicle only drives on the right side of the road.

\clearpage
\section{Execution Time Analysis}
\label{app:time-analysis}

\Cref{fig:execution-time-analysis} shows the execution time in seconds as a function of the number of simulation runs. The measurements were taken for the scripts used in the evaluation presented in \cref{sec:evaluation}, covering the exhaustive and responsibility-guided approaches in the sinusoidal short narrow environment. Importantly, these results demonstrate that the cost of all additional computation in the responsibility-guided approach is negligible relative to the number of simulation runs and therefore does not materially affect the overall runtime. We observe a small runtime overhead for the distance-based approaches, likely due to the distance function computations. However, this gap remains small and does not increase with the number of simulation runs and obstacles. Thus, the analysis in \cref{sec:evaluation} focuses on the number of simulation runs.

\pgfplotsset{
    exhaustiveDots/.style={
        color=black!70,
        only marks,
        mark=*,
        mark size=0.5pt
    },
    exhaustiveLine/.style={
        color=black!70,
        line width=0.6pt,
        no marks
    },
    discreteDots/.style={
        color=red!100,
        only marks,
        mark=*,
        mark size=0.5pt
    },
    discreteLine/.style={
        color=red!100,
        line width=0.6pt,
        no marks
    },
    euclideanDots/.style={
        color=blue!100,
        only marks,
        mark=*,
        mark size=0.5pt
    },
    euclideanLine/.style={
        color=blue!100,
        line width=0.6pt,
        no marks
    },
    simulationDots/.style={
        color=green!60!black,
        only marks,
        mark=*,
        mark size=0.5pt
    },
    simulationLine/.style={
        color=green!60!black,
        line width=0.6pt,
        no marks
    }
}

\begin{figure}[h]
    \centering

    \begin{tikzpicture}

        \begin{axis}[
            width=\columnwidth,
            height=0.55\columnwidth,
            ymin=0,
            xlabel={Simulation runs},
            ylabel={Execution time (sec.)},
            grid=major,
            grid style={black!10},
            axis line style={black!60},
            tick align=outside,
            tick pos=left,
            tick label style={
                font=\scriptsize
            },
            label style={
                font=\small
            },
            legend style={
                font=\scriptsize,
                draw=none,
                at={(1.00,0.00)},
                anchor=south east,
                legend columns=1
            }
        ]


        \addplot[
            exhaustiveDots,
            forget plot
        ] table[
            x=simulation_executions,
            y=execution_time_seconds,
            col sep=comma
        ] {results/exhaustive_observed_execution_time_analysis.csv};

        \addplot[
            exhaustiveLine,
            forget plot
        ] table[
            x=line_x,
            y=line_y,
            col sep=comma
        ] {results/exhaustive_observed_execution_time_trend_line.csv};


        \addplot[
            discreteDots,
            forget plot
        ] table[
            x=simulation_executions,
            y=execution_time_seconds,
            col sep=comma
        ] {results/responsibility_discrete_execution_time_analysis.csv};

        \addplot[
            discreteLine,
            forget plot
        ] table[
            x=line_x,
            y=line_y,
            col sep=comma
        ] {results/responsibility_discrete_execution_time_trend_line.csv};


        \addplot[
            euclideanDots,
            forget plot
        ] table[
            x=simulation_executions,
            y=execution_time_seconds,
            col sep=comma
        ] {results/responsibility_euclidean_execution_time_analysis.csv};

        \addplot[
            euclideanLine,
            forget plot
        ] table[
            x=line_x,
            y=line_y,
            col sep=comma
        ] {results/responsibility_euclidean_execution_time_trend_line.csv};


        \addplot[
            simulationDots,
            forget plot
        ] table[
            x=simulation_executions,
            y=execution_time_seconds,
            col sep=comma
        ] {results/responsibility_simulation_execution_time_analysis.csv};

        \addplot[
            simulationLine,
            forget plot
        ] table[
            x=line_x,
            y=line_y,
            col sep=comma
        ] {results/responsibility_simulation_execution_time_trend_line.csv};


        \addlegendimage{exhaustiveLine}
        \addlegendentry{Exhaustive}

        \addlegendimage{discreteLine}
        \addlegendentry{Discrete}

        \addlegendimage{euclideanLine}
        \addlegendentry{Dist. Euclidean}

        \addlegendimage{simulationLine}
        \addlegendentry{Dist. Chamfer}

        \end{axis}

    \end{tikzpicture}

    \caption{%
        Execution time as a function of the number of simulation runs.
    }
    \label{fig:execution-time-analysis}

\end{figure}
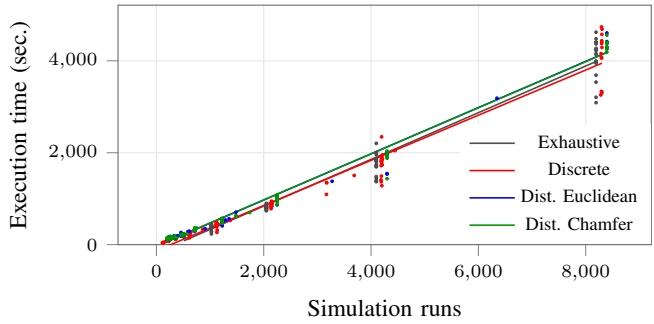

\clearpage
\section{Evaluation Results with Chamfer Distance}
\label{app:chamfer-distance-results}

\cref{fig:runs-comparison-chamfer}, \cref{tab:auc-results-chamfer}, and \cref{fig:responsibility-coverage-chamfer} report the evaluation results for the Chamfer distance. 

\pgfplotsset{
    exhaustive/.style={
        color=black!70,
        very thick,
        mark=*,
        mark size=1.3pt
    },
    discrete50/.style={
        color=red!50,
        thick,
        mark=square*,
        mark size=1.3pt
    },
    discrete100/.style={
        color=red!100,
        thick,
        mark=square*,
        mark size=1.3pt
    },
    euclidean50/.style={
        color=blue!50,
        thick,
        mark=triangle*,
        mark size=1.5pt
    },
    euclidean100/.style={
        color=blue!100,
        thick,
        mark=triangle*,
        mark size=1.5pt
    }
}











\begin{figure*}[t]
    \centering

    \begin{tikzpicture}

        \begin{groupplot}[
            group style={
                group name=plots,
                group size=6 by 3,
                horizontal sep=2mm,
                vertical sep=4mm,
            },
            width=0.238\textwidth,
            height=0.185\textwidth,
            xmin=9.7,
            xmax=14.3,
            ymin=0,
            ymax=18000,
            xtick={10,11,12,13,14},
            ytick={0,5000,10000,15000},
            yticklabels={0,5k,10k,15k},
            scaled y ticks=false,
            tick align=outside,
            tick pos=left,
            grid=major,
            grid style={black!10},
            axis line style={black!60},
            tick label style={font=\scriptsize},
            label style={font=\small},
            title style={font=\small},
        ]


        \nextgroupplot[
            title={Short Narrow},
            ylabel={Runs},
            xticklabels=\empty,
            legend columns=5,
            legend to name=commonlegendchamfer,
            legend style={
                font=\small,
                draw=none,
                /tikz/every even column/.append style={
                    column sep=3mm
                }
            }
        ]
        \plotrunswithlegend{results/chamfer/sinusoidal_short_narrow_runs_results.csv}

        \nextgroupplot[
            title={Short Wide},
            xticklabels=\empty,
            yticklabels=\empty
        ]
        \plotruns{results/chamfer/sinusoidal_short_wide_runs_results.csv}

        \nextgroupplot[
            title={Long Narrow},
            xticklabels=\empty,
            yticklabels=\empty
        ]
        \plotruns{results/chamfer/sinusoidal_long_narrow_runs_results.csv}

        \nextgroupplot[
            title={Long Wide},
            xticklabels=\empty,
            yticklabels=\empty
        ]
        \plotruns{results/chamfer/sinusoidal_long_wide_runs_results.csv}

        \nextgroupplot[
            hide axis,
            width=0.19\textwidth
        ]

        \nextgroupplot[
            ylabel={Runs},
            xticklabels=\empty
        ]
        \plotruns{results/chamfer/metadrive_curve_runs_results.csv}


        \nextgroupplot[
            ylabel={Runs},
            xticklabels=\empty
        ]
        \plotruns{results/chamfer/segmented_short_narrow_runs_results.csv}

        \nextgroupplot[
            xticklabels=\empty,
            yticklabels=\empty
        ]
        \plotruns{results/chamfer/segmented_short_wide_runs_results.csv}

        \nextgroupplot[
            xticklabels=\empty,
            yticklabels=\empty
        ]
        \plotruns{results/chamfer/segmented_long_narrow_runs_results.csv}

        \nextgroupplot[
            xticklabels=\empty,
            yticklabels=\empty
        ]
        \plotruns{results/chamfer/segmented_long_wide_runs_results.csv}

        \nextgroupplot[
            hide axis,
            width=0.19\textwidth
        ]

        \nextgroupplot[
            ylabel={Runs},
            xticklabels=\empty
        ]
        \plotruns{results/chamfer/metadrive_straight_runs_results.csv}


        \nextgroupplot[
            ylabel={Runs}
        ]
        \plotruns{results/chamfer/lane_shift_short_narrow_runs_results.csv}

        \nextgroupplot[
            yticklabels=\empty
        ]
        \plotruns{results/chamfer/lane_shift_short_wide_runs_results.csv}

        \nextgroupplot[
            yticklabels=\empty
        ]
        \plotruns{results/chamfer/lane_shift_long_narrow_runs_results.csv}

        \nextgroupplot[
            yticklabels=\empty
        ]
        \plotruns{results/chamfer/lane_shift_long_wide_runs_results.csv}

        \nextgroupplot[
            hide axis,
            width=0.19\textwidth
        ]

        \nextgroupplot[
            ylabel={Runs},
            xtick={10,11,12,13,14},
            xticklabels={10,11,12,13,14}
        ]
        \plotruns{results/chamfer/metadrive_mixed_runs_results.csv}

        \end{groupplot}


        \node[
            anchor=south,
            font=\small
        ] at ([yshift=7mm]
             $(plots c1r1.north)!0.5!(plots c4r1.north)$)
        {LiteRacer};

        \node[
            anchor=south,
            font=\small
        ] at ([yshift=7mm]
             $(plots c6r1.north)$)
        {MetaDrive};


        \node[
            rotate=90,
            anchor=south,
            font=\small
        ] at ([xshift=-12mm]$(plots c1r1.west)$)
        {Sinusoidal};

        \node[
            rotate=90,
            anchor=south,
            font=\small
        ] at ([xshift=-12mm]$(plots c1r2.west)$)
        {Segmented};

        \node[
            rotate=90,
            anchor=south,
            font=\small
        ] at ([xshift=-12mm]$(plots c1r3.west)$)
        {Lane Shift};


        \node[
            rotate=90,
            anchor=south,
            font=\small
        ] at ([xshift=-12mm]$(plots c6r1.west)$)
        {Curve};

        \node[
            rotate=90,
            anchor=south,
            font=\small
        ] at ([xshift=-12mm]$(plots c6r2.west)$)
        {Straight};

        \node[
            rotate=90,
            anchor=south,
            font=\small
        ] at ([xshift=-12mm]$(plots c6r3.west)$)
        {Mixed};


        \node[
            anchor=north
        ] at ([yshift=-5mm]
             $(plots c1r3.south)!0.5!(plots c6r3.south)$)
        {\pgfplotslegendfromname{commonlegendchamfer}};

    \end{tikzpicture}

    \caption{%
        Efficiency of explanation generation algorithms as a function of the number of obstacles ($10$--$14$), measured by the number of simulation runs (lower is better).
    }
    \label{fig:runs-comparison-chamfer}

\end{figure*}

\begin{table}[h]
    \scriptsize
    \centering

    \caption{%
        AUC results for the discrete and distance-based rankings
        across different environments and numbers of obstacles.
    }
    \label{tab:auc-results-chamfer}

    \scriptsize
    \setlength{\tabcolsep}{2.5pt}

    \begin{tabular}{c c || cc cc cc cc cc}

        \toprule
        \multicolumn{2}{c}{} &
        \multicolumn{2}{c}{\textbf{10}} &
        \multicolumn{2}{c}{\textbf{11}} &
        \multicolumn{2}{c}{\textbf{12}} &
        \multicolumn{2}{c}{\textbf{13}} &
        \multicolumn{2}{c}{\textbf{14}}
        \\
        \cmidrule(lr){3-4}\cmidrule(lr){5-6}\cmidrule(lr){7-8}\cmidrule(lr){9-10}\cmidrule(lr){11-12}
        & \textbf{Env.} &
        \textbf{Disc.} & \textbf{Dist.} &
        \textbf{Disc.} & \textbf{Dist.} &
        \textbf{Disc.} & \textbf{Dist.} &
        \textbf{Disc.} & \textbf{Dist.} &
        \textbf{Disc.} & \textbf{Dist.}
        \\
        \midrule


        \multirow{12}{*}{\rotatebox[origin=c]{90}{LiteRacer}}
        & Sin-SN
        & 0.21 & \textbf{0.29}
        & 0.15 & \textbf{0.35}
        & 0.10 & \textbf{0.45}
        & 0.19 & \textbf{0.41}
        & 0.14 & \textbf{0.30}
        \\

        & Sin-SW
        & 0.19 & \textbf{0.35}
        & 0.17 & \textbf{0.31}
        & 0.12 & \textbf{0.32}
        & 0.12 & \textbf{0.26}
        & 0.13 & \textbf{0.32}
        \\

        & Sin-LN
        & 0.21 & \textbf{0.36}
        & 0.17 & \textbf{0.33}
        & 0.20 & \textbf{0.38}
        & 0.18 & \textbf{0.33}
        & 0.08 & \textbf{0.30}
        \\

        & Sin-LW
        & 0.13 & \textbf{0.34}
        & 0.15 & \textbf{0.38}
        & 0.14 & \textbf{0.35}
        & 0.10 & \textbf{0.24}
        & 0.17 & \textbf{0.36}
        \\
        \cmidrule(lr){2-12}

        & Seg-SN
        & 0.19 & \textbf{0.39}
        & 0.20 & \textbf{0.34}
        & 0.13 & \textbf{0.44}
        & 0.19 & \textbf{0.40}
        & 0.17 & \textbf{0.37}
        \\

        & Seg-SW
        & 0.21 & \textbf{0.35}
        & 0.18 & \textbf{0.33}
        & 0.15 & \textbf{0.30}
        & 0.14 & \textbf{0.29}
        & 0.11 & \textbf{0.30}
        \\

        & Seg-LN
        & 0.22 & \textbf{0.44}
        & 0.20 & \textbf{0.34}
        & 0.19 & \textbf{0.37}
        & 0.12 & \textbf{0.39}
        & 0.11 & \textbf{0.35}
        \\

        & Seg-LW
        & 0.13 & \textbf{0.32}
        & 0.16 & \textbf{0.30}
        & 0.11 & \textbf{0.29}
        & 0.12 & \textbf{0.26}
        & 0.10 & \textbf{0.24}
        \\
        \cmidrule(lr){2-12}

        & LS-SN
        & 0.23 & \textbf{0.41}
        & 0.19 & \textbf{0.40}
        & 0.14 & \textbf{0.46}
        & 0.12 & \textbf{0.28}
        & 0.17 & \textbf{0.36}
        \\

        & LS-SW
        & 0.18 & \textbf{0.22}
        & 0.11 & \textbf{0.19}
        & 0.13 & \textbf{0.34}
        & 0.11 & \textbf{0.29}
        & 0.12 & \textbf{0.27}
        \\

        & LS-LN
        & 0.19 & \textbf{0.39}
        & 0.17 & \textbf{0.36}
        & 0.17 & \textbf{0.41}
        & 0.20 & \textbf{0.39}
        & 0.20 & \textbf{0.39}
        \\

        & LS-LW
        & 0.15 & \textbf{0.37}
        & 0.12 & \textbf{0.33}
        & 0.15 & \textbf{0.29}
        & 0.13 & \textbf{0.33}
        & 0.16 & \textbf{0.41}
        \\
        \cmidrule(lr){2-12}


        \multirow{3}{*}{\rotatebox[origin=c]{90}{MetaDr.}}
        & Curve
        & 0.34 & \textbf{0.43}
        & 0.28 & \textbf{0.48}
        & 0.20 & \textbf{0.42}
        & 0.15 & \textbf{0.34}
        & 0.16 & \textbf{0.56}
        \\

        & Straight
        & \textbf{0.41} & 0.40
        & 0.28 & \textbf{0.31}
        & 0.22 & \textbf{0.35}
        & 0.20 & \textbf{0.36}
        & 0.25 & \textbf{0.32}
        \\

        & Mixed
        & 0.32 & \textbf{0.40}
        & 0.25 & \textbf{0.36}
        & 0.20 & \textbf{0.41}
        & 0.16 & \textbf{0.37}
        & 0.22 & \textbf{0.39}
        \\
        \bottomrule
    \end{tabular}

\end{table}

\input{chamfer_results/multiple_explanations_plot}

\clearpage
\section{Multiple Explanation Evaluation Euclidean Distance}
\label{app:euclidean-distance-results}

\cref{fig:responsibility-coverage-all} reports the ranking curves of all settings for the Euclidean distance-weighted responsibility.

\input{multiple_explanations_plot_all}
\end{appendices}


\end{document}